\documentclass[11pt]{article}
\usepackage{enumerate}
\usepackage[OT1]{fontenc}
\usepackage{multirow}
\usepackage{amsmath,amssymb}
\usepackage{natbib}
\usepackage[usenames]{color}
\usepackage{dsfont,makecell,subfigure}
\usepackage{smile}
\usepackage{multirow}

\usepackage[colorlinks,
            linkcolor=red,
            anchorcolor=blue,
            citecolor=blue
            ]{hyperref}

\usepackage{mathrsfs}

\usepackage{fullpage}

\usepackage{hyperref}
\usepackage[    protrusion=true,
            expansion=true,
            final,
            babel
                ]{microtype}
\def\##1\#{\begin{align}#1\end{align}}
\def\$#1\${\begin{align*}#1\end{align*}}
\usepackage{enumitem}

\begin{document}
\title{ Sparse Generalized Eigenvalue Problem: Optimal Statistical Rates via Truncated Rayleigh Flow}
\author{Kean Ming Tan, Zhaoran Wang, Han Liu, and Tong Zhang}
\maketitle

\begin{abstract}
\vskip4pt
Sparse generalized eigenvalue problem (GEP) plays a pivotal role in a large family of high-dimensional statistical models, including sparse Fisher's discriminant analysis,  canonical correlation analysis, and  sufficient dimension reduction. Sparse GEP involves solving a non-convex optimization problem.  Most existing methods and theory in the context of specific statistical models that are special cases of the sparse GEP require restrictive structural assumptions on the input matrices. In this paper, we propose a two-stage computational framework to solve the sparse GEP.  At the first stage, we solve a convex relaxation of the sparse GEP.  Taking the solution as an initial value, we then exploit a nonconvex optimization perspective and propose the truncated \underline{R}ayle\underline{i}gh \underline{fl}ow m\underline{e}thod (Rifle) to estimate the leading generalized eigenvector. We show that Rifle converges linearly to a solution with the optimal statistical rate of convergence for many statistical models. Theoretically, our method significantly improves upon the existing literature by eliminating structural assumptions on the input matrices for both stages.  To achieve this, our analysis involves two key ingredients: (i) a new analysis of the gradient based method on nonconvex objective functions, and (ii) a fine-grained characterization of the evolution of sparsity patterns along the solution path. Thorough numerical studies are provided to validate the theoretical results. \\


\end{abstract}

\section{Introduction}
\label{section:introduction}
A large class of high-dimensional statistical methods such as canonical correlation analysis (CCA),  Fisher's discriminant analysis (FDA), and sufficient dimension reduction (SDR) can be formulated as the generalized eigenvalue problem (GEP). Let $\Ab\in\RR^{d\times d}$ be a symmetric matrix and let $\Bb \in \RR^{d\times d}$ be a  positive definite matrix. For a symmetric-definite matrix pair~$(\Ab,\Bb)$, the generalized eigenvalue problem aims to obtain $\vb^*\in \RR^d$ satisfying
\#
\label{Eq:gep opt}
 \Ab \vb^* = \lambda_{\max} (\Ab,\Bb) \cdot \Bb \vb^*,
\#
where  $\vb^*$ is the leading generalized eigenvector corresponding to the largest generalized eigenvalue $\lambda_{\max} (\Ab,\Bb)$ of the matrix pair $(\Ab, \Bb)$. 
The largest generalized eigenvalue can also be characterized as 
\[
\lambda_{\max} (\Ab,\Bb) = \max_{\vb\in \RR^{d}} \vb^T\Ab \vb, \qquad \mathrm{subject~to~} \vb^T \Bb \vb = 1. 
\]

In many real-world applications, the matrix pair $(\Ab,\Bb)$ is a population quantity that is unknown in general. Instead, we can only access $(\hat{\Ab},\hat{\Bb})$, which is an estimator of  $ (\Ab,\Bb)$ on the basis of $n$ independent observations: 
\[
\hat{\Ab} = \Ab + \Eb_{\Ab} \quad \mathrm{and}\quad \hat{\Bb} = \Bb + \Eb_{\Bb},
\]
where $\Eb_{\Ab}$ and $\Eb_{\Bb}$  are stochastic errors due to finite sample estimation. For  statistical models considered in this paper, $\Eb_{\Ab}$ and $\Eb_{\Bb}$ are symmetric matrices. 

In the high-dimensional setting in which $d > n$, we assume that the leading generalized eigenvector $\vb^*$ is sparse.
Let $s=\|\vb^* \|_0$ be the  number of non-zero entries in $\vb^*$, and assume that $s$ is much smaller than $n$ and $d$.
We aim to estimate $\vb^*$ based on $\hat{\Ab}$ and $\hat{\Bb}$ by  solving the following optimization problem  
\#
\label{Eq:gep opt}
\underset{\vb \in \RR^{d}}{\mathrm{maximize}}~~ \vb^T \hat{\Ab} \vb, \quad \mathrm{subject~to~} 
\vb^T \hat{\Bb} \vb = 1,\quad \|\vb\|_0 \le s. 
\#
There are three major challenges in solving \eqref{Eq:gep opt}. Firstly,  in the high-dimensional setting, $\hat{\Bb}$ is singular and not invertible, and classical algorithms which require taking the inverse of $\hat{\Bb}$  are not directly applicable \citep{golub2012matrix}.
Secondly, due to the normalization term $\vb^T \hat{\Bb} \vb = 1$, many recent proposals for solving sparse eigenvalue problem such as the truncated power method in \citet{yuan2013truncated} cannot be directly applied to solve (\ref{Eq:gep opt}).
Thirdly, \eqref{Eq:gep opt} requires maximizing a convex objective function over a nonconvex set, which is  NP-hard even when $\hat{\Bb}$ is the identity matrix \citep{moghaddam2006generalized, moghaddam2006spectral}.  

In this paper, we propose a two-stage computational framework for solving the sparse GEP in \eqref{Eq:gep opt}.  At the first stage, we solve a convex relaxation of \eqref{Eq:gep opt}.  Our proposal  generalizes the convex relaxation proposed in \citet{gao2014sparse} in the context of sparse CCA to the sparse GEP setting.  
\citet{gao2014sparse} assumes that $\Ab$ is low rank, positive semidefinite, and the rank of $\Ab$ is known.  Our theoretical analysis removes all of the aforementioned assumptions.
Using the solution as an initial value, we propose a nonconvex optimization algorithm to solve (\ref{Eq:gep opt}) directly. The proposed algorithm iteratively performs a gradient ascent step on the generalized Rayleigh quotient $\vb^T \hat{\Ab} \vb/\vb^T \hat{\Bb} \vb$, and a truncation step that preserves the top $k$ entries  of $\vb$ with the largest magnitudes while setting the remaining entries to zero. Here, 
 $k$ is a tuning parameter that controls the cardinality of the solution. 
 Theoretical guarantees are established for the proposed nonconvex algorithm.
To the best of our knowledge, this is the first general theoretical result for sparse generalized eigenvalue problem in the high-dimensional setting.  

We provide a brief description of the theoretical result for the nonconvex algorithm at the second stage.  
Let $\{\vb_t\}_{t=0}^L$ be the solution sequence resulting from the proposed algorithm, where $L$ is the total number of iterations and $\vb_0$ is the initialization point. We prove that, under~mild conditions, 
\#\label{eq:w1}
\| \vb^t - \vb^* \|_2 \leq \underbrace{\nu^t \cdot \|\vb^0 - \vb^*\|_2}_{\displaystyle\rm optimization~error} + \underbrace{\frac{\sqrt{\rho(\Eb_{\Ab}, 2k+s)^2 + \rho(\Eb_{\Bb}, 2k+s)^2}}{\xi(\Ab, \Bb)}}_{\displaystyle\rm statistical~error}\quad (t = 1,\ldots, L).
\#
The quantities $\nu \in (0,1)$ and $\xi(\Ab,\Bb)$ depend on the population matrix pair $(\Ab, \Bb)$.  These quantities will be specified in Section~\ref{section:theory}. Meanwhile, $\rho(\Eb_{\Ab}, 2k+s)$ is defined as 
\#\label{eq:w3}
\rho(\Eb_{\Ab}, 2k+s) = \sup_{\|\ub\|_2 = 1, \|\ub\|_0 \leq 2k+s} | \ub^T \Eb_{\Ab} \ub |
\#
and $\rho(\Eb_{\Bb}, 2k+s)$ is defined similarly. The first term on the right-hand side quantifies the exponential decay of the optimization error, while the second term characterizes the statistical error due to  finite  sample estimation. 
In particular, for many statistical models that can be formulated as a sparse GEP such as  sparse CCA, sparse FDA, and sparse SDR, we establish that 
\#\label{eq:w2}
\max\{\rho(\Eb_{\Ab}, 2k+s), \rho(\Eb_{\Bb}, 2k+s)\} \leq \sqrt{\frac{(s+2k) \log d}{n}}
\#
with high probability. Consequently, for any properly chosen $k$ that is of the same order as $s$, the algorithm achieves an estimator of $\vb^*$ with the optimal statistical rate of convergence $\sqrt{s \log d/n}$.

The sparse generalized eigenvalue problem in \eqref{Eq:gep opt} is also closely related to the classical matrix computation literature (see, e.g., \citealp{golub2012matrix} for a survey, and more recent results in \citealp{ge2016efficient}). There are two key differences between our results and existing work. Firstly, we have an additional nonconvex constraint on the sparsity level, which allows us to handle the high-dimensional setting. Secondly, due to the existence of stochastic errors, we allow the normalization matrix $\hat{\Bb}$ to be rank-deficient, while in the classical setting $\hat{\Bb}$ is assumed to be positive definite. In comparison with  existing generalized eigenvalue algorithms, our algorithm keeps the iterative solution sequence within a basin that involves only a few coordinates of $\vb$ such that the corresponding submatrix of $\hat{\Bb}$ is positive definite. Moreover, our algorithm ensures that the statistical errors \eqref{eq:w1} are in terms of the largest sparse eigenvalues of the stochastic errors $\Eb_{\Ab}$ and $\Eb_{\Bb}$, which is defined in \eqref{eq:w3}. In contrast, a straightforward application of the classical matrix perturbation theory gives  statistical error terms that involve the largest eigenvalues of $\Eb_{\Ab}$ and $\Eb_{\Bb}$, which are much larger than their corresponding sparse eigenvalues \citep{stewart1990}.  

An \texttt{R} package for fitting the sparse generalized eigenvalue problem will be uploaded to \texttt{CRAN}.\\


\noindent \textbf{Notation:} 
Let $\vb = (v_1,\ldots,v_d)^T \in \RR^d$. We define the $\ell_q$-norm of $\vb$ as $\|\vb\|_q = (\sum_{j=1}^d |v_j|^q)^{1/q}$~for $1\le q < \infty$. Let $\lambda_{\max}(\Zb)$ and $\lambda_{\min}(\Zb)$ be the largest and smallest eigenvalues correspondingly. 
If $\Zb$ is positive definite, we define its condition number as $\kappa(\Zb) = \lambda_{\max}(\Zb)/\lambda_{\min} (\Zb)$. We denote  $\lambda_k(\Zb)$ to be the $k$th eigenvalue of $\Zb$, and the spectral norm of $\Zb$ by $\|\Zb\|_2  = \sup_{\|\vb\|_2 =1}\; \|\Zb \vb \|_2$. Furthermore, let $\|\Zb\|_{1,1}=\sum_{i,j} |Z_{ij}|$, $\|\Zb\|_{\infty,\infty} = \underset{i,j}{\max}~|Z_{ij}|$ and $\|\Zb\|_* = \mathrm{tr}(\Zb)$.  For $F\subset \{1,\ldots,d\}$, 
 let $\Zb_{\cdot F} \in \RR^{d\times |F|}$ and $\Zb_{F\cdot } \in \RR^{ |F| \times d}$  be the submatrix of $\Zb$ where the columns and rows are restricted to the set $F$, respectively.
With some abuse of notation, let $\Zb_F\in \RR^{|F|\times |F|}$ be the submatrix of $\Zb$, where the rows and columns are restricted to the set $F$. 
In addition,  Finally, we define $\rho(\Zb, s) = \sup_{\|\ub\|_2 = 1, \|\ub\|_0 \leq s} | \ub^T \Zb \ub |$.

\section{Sparse Generalized Eigenvalue Problem and Its Applications}
\label{section:previous work}

Many high-dimensional multivariate statistics methods can be formulated as special instances of (\ref{Eq:gep opt}).
For instance, when $\hat{\Bb} = \Ib$, (\ref{Eq:gep opt}) reduces to the sparse principal component analysis (PCA) that has received considerable attention within the past decade (among others, \citealp{zou2006sparse,d2007direct,d2008optimal,witten2009penalized,ma2013sparse,cai2013sparse,yuan2013truncated,vu2013fantope,vu2013minimax,birnbaum2013minimax,wang2013sparse,wang2014tighten, gu2014sparse}).
In the following, we provide three examples when $\hat{\Bb}$ is not the identity matrix. 
We start with sparse Fisher's discriminant analysis for classification problem   (among others, \citealp{tibshirani2003class,guo2007regularized,leng2008sparse,clemmensen2012sparse,mai2012direct,mai2015multiclass,kolar2015optimal,gaynanova2015optimal,fan2015quadro}).

\begin{example}
\label{example:fda}
\textbf{Sparse Fisher's discriminant analysis:} Given $n$ observations with $K$ distinct classes, Fisher's discriminant problem seeks a low-dimensional projection of the observations such that the between-class variance, $\bSigma_b$, is large relative to the within-class variance, $\bSigma_w$.    Let  $\hat{\bSigma}_b$ and $\hat{\bSigma}_w$ be estimators of $\bSigma_b$ and $\bSigma_w$, respectively.  To obtain a sparse leading discriminant vector, one solves
\begin{equation}
\label{Eq:FDA}
\underset{\vb}{\mathrm{maximize}} \; \vb^T \hat{\bSigma}_b \vb, \qquad \mathrm{subject\; to\;} 
\vb^T \hat{\bSigma}_w \vb = 1, \qquad \|\vb\|_0 \le s.
\end{equation}
This is a special case of (\ref{Eq:gep opt}) with $\hat{\Ab} = \hat{\bSigma}_b$ and $\hat{\Bb}= \hat{\bSigma}_w$.  

\end{example}

Next, we consider sparse canonical correlation analysis that explores the relationship between two high-dimensional random vectors \citep{witten2009penalized, chen2013sparse,gao2014sparse,gao2015minimax}.   
\begin{example}
\label{example:cca}
\textbf{Sparse canonical correlation analysis:} 
  Let $\bX$ and $\bY$ be two  random vectors.   Let $\bSigma_{x}$ and  $\bSigma_{y}$ be the covariance matrices for $\bX$ and  $\bY$, respectively, and let $\bSigma_{xy}$ be the cross-covariance matrix between $\bX$ and $\bY$. 
To obtain sparse leading canonical direction vectors, we solve
\begin{equation}
\label{Eq:CCA}
\underset{\vb_x,\vb_y}{\mathrm{maximize}} \; \vb_x^T \hat{\bSigma}_{xy} \vb_y, \qquad \mathrm{subject\; to\;} 
\vb^T_x \hat{\bSigma}_x \vb_x = \vb^T_y \hat{\bSigma}_y \vb_y = 1, \quad \|\vb_x\|_0 \le s_x, \quad \|\vb_y\|_0 \le s_y,
\end{equation}
where $s_x$ and $s_y$ control the cardinality of $\vb_x$ and $\vb_y$.
This is a special case of (\ref{Eq:gep opt}) with 
\[
\hat{\Ab}= \begin{pmatrix}  \mathbf{0} & \hat{\bSigma}_{xy} \\ 
\hat{\bSigma}_{xy} & \mathbf{0}  \end{pmatrix}, 
\qquad
\hat{\Bb}= \begin{pmatrix}  \hat{\bSigma}_{x} &\mathbf{0} \\ 
 \mathbf{0} & \hat{\bSigma}_{y}  \end{pmatrix},
\qquad 
\vb = \begin{pmatrix} \vb_x \\ \vb_y
\end{pmatrix}.
\]
\end{example}
\noindent Theoretical guarantees for sparse CCA were established recently.  \citet{chen2013sparse} proposed a nonconvex optimization algorithm for solving (\ref{Eq:CCA}) with theoretical guarantees. However, their algorithm involves obtaining accurate estimators of $\bSigma_x^{-1}$ and $\bSigma_y^{-1}$, which are in general difficult to obtain without imposing sparsity assumption on $\bSigma_x^{-1}$ and $\bSigma_y^{-1}$. 
In a follow-up work, \citet{gao2014sparse} proposed a two-stage procedure that attains the optimal statistical rate of convergence \citep{gao2015minimax}.
However, they require the matrix $\bSigma_{xy}$ to be low-rank, positive semidefinite, and that the rank of $\bSigma_{xy}$ is known a priori.   As suggested in \citet{gao2015minimax}, the low-rank assumption on $\bSigma_{xy}$ may be unrealistic in many real data applications where one is interested in recovering the first few sparse canonical correlation directions while there might be additional directions in the population structure. 
Our proposal does not impose any structural assumption on $\bSigma_x$, $\bSigma_y$, and we only require $\bSigma_{xy}$ to be approximately low rank in the sense that the leading generalized eigenvalue is larger than the remaining. 

Next, we consider a regression problem with a univariate response $Y$ and $d$-dimensional covariates $\bX$, with the goal of inferring the conditional distribution of $Y$ given $\bX$. 
Sufficient dimension reduction is a popular approach for reducing the dimensionality of the covariates \citep{li1991sliced,cook1999dimension,dennis2000save,cook2007fisher,cook2008principal,ma2013review}. 
It can be shown that many sufficient dimension reduction methods can be formulated as generalized eigenvalue problems \citep{li2007sparse,chen2010coordinate}. 
In the following, we consider the sparse sliced inverse regression \citep{li1991sliced}. 
   
\begin{example}
\label{example:sir}
\textbf{Sparse sliced inverse regression:} 
Consider the model
\[
Y= f(\vb_1^T \bX,\ldots,\vb_K^T \bX,\epsilon),
\]
where $\epsilon$ is the stochastic error independent of $\bX$, and $f(\cdot)$ is an unknown link function. \citet{li1991sliced} proved that under regularity conditions, the subspace spanned by $\vb_1,\ldots,\vb_K$ can be identified.
Let $\bSigma_x$ be the covariance matrix for $\bX$ and let $\bSigma_{E(\bX\mid Y)}$ be the covariance matrix of the conditional expectation $E(\bX\mid Y)$.
The first leading eigenvector of the subspace spanned by $\vb_1,\ldots,\vb_K$ can be identified by solving
\begin{equation}
\label{Eq:sir}
\underset{\vb}{\mathrm{maximize}} \; \vb^T \hat{\bSigma}_{E(\bX\mid Y)}\vb, \qquad \mathrm{subject\; to\;} 
\vb^T \hat{\bSigma}_x \vb = 1, \qquad \|\vb\|_0 \le s.
\end{equation}
This is a special case of (\ref{Eq:gep opt}) with $\hat{\Ab}= \hat{\bSigma}_{E(\bX\mid Y)}$ and $\hat{\Bb} = \hat{\bSigma}_{x}$.
\end{example}
Many authors have proposed methods for sparse sliced inverse regression \citep{li2006sparse,zhu2006sliced,li2008sliced,chen2010coordinate,yin2015sequential}.  
More generally, in the context of sparse sufficient dimension reduction, \citet{li2007sparse} and \citet{chen2010coordinate}~reformulated sparse sufficient dimension reduction problems into the sparse generalized eigenvalue problem in (\ref{Eq:gep opt}).  However, these approaches lack algorithmic and non-asymptotic statistical guarantees in the high-dimensional setting. Our results are applicable to most sparse sufficient dimension reduction methods.

\section{Methodology and Algorithm}
\label{section:algorithm}
In Section~\ref{sec:tgd}, we propose an iterative algorithm to estimate $\vb^*$ by solving (\ref{Eq:gep opt}), which we refer to as truncated \underline{R}ayle\underline{i}gh \underline{fl}ow m\underline{e}thod (Rifle).  
Rifle requires an input of an initial vector $\mathbf{v}_0$ that is sufficiently close to $\vb^*$.  To this end, we propose a convex optimization approach to obtain such an initial vector $\mathbf{v}_0$ in Section~\ref{convexrelax}.  

\subsection{Truncated Rayleigh Flow Method (Rifle)}
\label{sec:tgd}
Optimization problem (\ref{Eq:gep opt}) can be rewritten as 
\[
\underset{\vb\in\RR^d}{\mathrm{maximize}} \; \frac{\vb^T \hat{\Ab} \vb}{\vb^T \hat{\Bb} \vb}, \qquad\mathrm{subject\; to\; } \|\vb\|_0 \le s,
\]
where the objective function is generally referred to as the generalized Rayleigh quotient.

The main crux of our proposed algorithm is as follows.
Given an initial vector $\mathbf{v}_0$, we first compute the gradient of the  generalized Rayleigh quotient. We then update the initial vector by its ascent direction and normalize it such that the updated vector has norm one.  
This step ensures that the generalized Rayleigh quotient for the updated vector is at least as large as that of the initial vector.
Indeed, in Theorem~\ref{theorem:main}, we show that if the initial vector $\mathbf{v}_0$ is close to $\mathbf{v}^*$, then this step ensures that the updated vector is closer to $\mathbf{v}^*$ compared to $\mathbf{v}_0$.
Next, we truncate the updated vector by keeping the elements with the largest $k$ absolute values and setting the remaining elements to zero.  This step ensures that the updated vector is $k$-sparse, i.e., only $k$ entries are non-zero.  
Finally, we normalize the updated vector such that it has norm one.  
These steps are repeated until convergence.
We summarize the details in Algorithm~\ref{alg:tgd}.
\begin{algorithm}[!htp]
\caption{\label{alg:tgd} Truncated \underline{R}ayle\underline{i}gh \underline{Fl}ow M\underline{e}thod (Rifle)}
 \textbf{Input}: matrices $\hat{\Ab}$, $\hat{\Bb}$,  initial vector $\vb_0$,  cardinality $k \in \{1,\ldots,d\}$, and step size $\eta$.\\
 \textbf{Truncate}: Truncate $\vb_0$ by keeping the largest $k$ absolute elements, and setting the remaining entries to zero.

Let $t=1$.  Repeat the following until convergence:
\begin{enumerate}
\item  $\rho_{t-1} \leftarrow  \vb_{t-1}^T \hat{\Ab} \vb_{t-1}/   \vb_{t-1}^T \hat{\Bb} \vb_{t-1}$.
\item $\Cb\leftarrow \Ib+ (\eta/\rho_{t-1})\cdot  (\hat{\Ab}-\rho_{t-1}\hat{\Bb}) $.
\item    $\vb_t' \leftarrow\Cb \vb_{t-1}/\|\Cb \vb_{t-1}\|_2$. 
\item Let $F_t = \mathrm{supp}(\vb_t' ,k)$ contain  the indices of $\vb_t'$ with the largest $k$ absolute values and   $\mathrm{Truncate}(\vb_t',F_t)$ be the truncated vector of $\vb_t'$ by setting $(\vb_t')_i = 0$ for $i\notin F_t$. 
\item  $\hat{\vb}_t \leftarrow \mathrm{Truncate}(\vb'_t,F_t)$.
\item  $\vb_t \leftarrow \hat{\vb}_t / \|\hat{\vb}_t\|_2$.  
\item $t \leftarrow t+1$.
\end{enumerate}
 \textbf{Output}: $\vb_t$.
\end{algorithm}

In addition to an initial vector $\vb_0$, Algorithm~\ref{alg:tgd} requires the choice of a step size $\eta$ and a tuning parameter $k$ on the cardinality of the solution.  As suggested by the theoretical results in Section~\ref{section:theory}, we need  $\eta$ to be sufficiently small such that $\eta \lambda_{\max} (\hat{\Bb})<1$.
In practice, the tuning parameter $k$ can be selected using cross-validation or based on prior knowledge. The computational complexity for each iteration of Algorithm~\ref{alg:tgd} is $\cO(kd+d)$: $\cO(d)$ for selecting the $k$ largest elements of a $d$-dimensional vector to obtain the set $F_t$, and $\cO(kd)$ for taking the product between a truncated vector and a matrix with columns restricted to the set $F_t$, and for calculating the difference between two matrices with columns restricted to the set $F_t$.

\subsection{A Convex Optimization Approach to Obtain $\mathbf{v}_0$}
\label{convexrelax}

As mentioned in Section~\ref{sec:tgd}, it is crucial to obtain an initial vector $\vb_0$ that is close to $\vb^*$ for Rifle.  
\citet{gao2014sparse} have proposed a convex formulation to estimate subspace spanned by the $K$ leading generalized eigenvectors for sparse CCA, under the assumption that $\Ab$ is low rank and positive semidefinite.  
Rather than estimating the $K$ leading generalized eigenvectors, the main idea of \citet{gao2014sparse} is to obtain an estimator of the subspace spanned by the $K$ leading generalized eigenvectors directly.  
In this section, we point out the fact that the proposed convex relaxation can be used more generally to estimate subspace of a sparse generalized eigenvalue problem, without the low rank and positive semidefinite structural assumptions on $\Ab$.

Similar to \eqref{Eq:gep opt}, the optimization problem for estimating the $K$ generalized eigenvectors can be written as 
\[
\underset{\Ub \in\RR^{d \times K}}{\mathrm{minimize}}~~ - \mathrm{tr}\left(\Ub^T \hat{\Ab}\Ub \right),\qquad \mathrm{subject~to~} \Ub^T\hat{\Bb} \Ub=\Ib_K.
\]
Rather than estimating the $K$ generalized eigenvectors which involves minimizing a concave function, we consider approximating the subspace spanned by these generalized eigenvectors.
Let $\Pb = \Ub \Ub^T$ and let $\cO= \{  \hat{\Bb}^{1/2} \Pb \hat{\Bb}^{1/2}   :\Ub^T \hat{\Bb} \Ub = \Ib_K\}$. 
By a change of variable, we obtain
\begin{equation}
\label{eq:convexrelaxation00}
\underset{\Pb\in \RR^{d\times d}}{\mathrm{minimize}}~~ -\mathrm{tr}\left(\hat{\Ab}\Pb \right),\qquad \mathrm{subject~to~} \Pb \in \cO,
\end{equation}
where the objective function is now linear in $\Pb$.

We consider the following convex relaxation of \eqref{eq:convexrelaxation00}, with a lasso penalty on $\Pb$ to encourage the estimated subspace to be sparse:
\begin{equation}
\label{eq:convexrelaxation}
\underset{\Pb\in \RR^{d\times d}}{\mathrm{minimize}}~ -\mathrm{tr}\left( \hat{\Ab} \Pb \right) + \zeta \|\Pb\|_{1,1}, \qquad \mathrm{subject~to}~ \|\hat{\Bb}^{1/2}\Pb \hat{\Bb}^{1/2}\|_* \le K~~\mathrm{and}~~ \|\hat{\Bb}^{1/2}\Pb \hat{\Bb}^{1/2}\|_2  \le 1,
\end{equation}
where $\|\cdot\|_*$ and $\|\cdot\|_2$ are the nuclear norm and spectral norm that encourage the solution to be low rank and that its eigenvalue is bounded, respectively.
Here, $\zeta$ and $K$ are two tuning parameters that encourages the estimated subspace $\Pb$ to be sparse and low rank, respectively.
The convex optimization problem \eqref{eq:convexrelaxation} can be solved using the alternating direction methods of multiplier algorithm, which we  summarize the details in Algorithm~\ref{alg:admm} \citep{BoydADMM,ADMMconvergence}.
The computational bottleneck in Algorithm~\ref{alg:admm} is the singular value decomposition on a $d\times d$ matrix, thus yielding a computational complexity of $\cO(d^3)$. Compare to the computational complexity of $\cO(kd+d)$ for Algorithm~\ref{alg:tgd}, it can be seen that obtaining a good initial vector $\vb_0$ is much more time consuming than refining the initial value.

Let $\hat{\Pb}$ be an estimator obtained from solving~\eqref{eq:convexrelaxation}.  
Then, the initial value $\vb_0$ can be set to be the largest eigenvector of $\hat{\Pb}$.
The theoretical guarantees for $\vb_0$ obtained via this approach are presented in Proposition~\ref{prop:initial} in Section~\ref{subsec:initialization}.
In practice, for the purpose of obtaining an initial value $\vb_0$, one can simply set $K=1$ and $\zeta$ to be approximately $\sqrt{\log d /n}$. In fact, we suggest setting $\zeta$ conservatively since there is a refinement step using Rifle to obtain an estimator that is closer to $\vb^*$.

\begin{algorithm}[!t]
\caption{\label{alg:admm} ADMM Algorithm for Solving \eqref{eq:convexrelaxation}}
 \textbf{Input}: matrices $\hat{\Ab}$, $\hat{\Bb}$, tuning parameters $\zeta$, $K$, ADMM parameter $\nu$, and convergence criterion $\epsilon$.  

 \textbf{Initialize}: matrices $\Pb_{0}$, $\Hb_{0}$, and ${\bGamma}_{0}$.

Let $t=1$.  Repeat the following until $\|{\Pb}_{t+1}-{\Pb}_{t}\|_F\le \epsilon$:
\begin{enumerate}
\item  Update $\Pb$ by solving the following lasso problem:
\[
{\Pb}_{t+1} = \underset{\Pb}{\argmin}~ \frac{\nu}{2} \|\hat{\Bb}^{1/2}\Pb\hat{\Bb}^{1/2} - \Hb_t + \bGamma_t    \|_{F}^2 - \mathrm{tr}(\hat{\Ab} \Pb) + \zeta \|\Pb\|_{1,1}.
\]
\item Let $\sum_{j=1}^d \omega_j \ab_j \mathbf{a}_j^T $ be the singular value decomposition of $\bGamma_{t} + \hat{\Bb}^{1/2} \Pb_{t+1}\hat{\Bb}^{1/2}$ and let 
\[
\gamma^* = \underset{\gamma>0}{\argmin}~\gamma\qquad \mathrm{subject~to}~ \sum_{j=1}^d 	\min \{1,\max(\omega_j-\gamma,0)\} \le K.
\]
Update $\Hb$ by 
\[
{\Hb}_{t+1} = \sum_{j=1}^d \min \{1,\max(\omega_j-\gamma^*,0)\}    \ab_j\mathbf{a}_j^T.
\]

\item   Update $\bGamma$ by 
\[
{\bGamma}_{t+1} = {\bGamma}_{t} + \hat{\Bb}^{1/2}{\Pb}_{t+1}\hat{\Bb}^{1/2} - {\Hb}_{t+1}.
\]
\item $t \leftarrow t+1$.
\end{enumerate}
\end{algorithm}

\section{Theoretical Results}
\label{section:theory}
We show that if the matrix pair $(\Ab,\Bb)$ has a unique sparse leading generalized eigenvector, then 
  Algorithm~\ref{alg:tgd} can accurately recover the population leading generalized eigenvector from the noisy matrix pair $(\hat{\Ab},\hat{\Bb})$.
Recall from the Introduction that  $\Ab$ is symmetric and $\Bb$ is positive definite.  
This condition ensures that all generalized eigenvalues are real. 
Recall that $\vb^*$ is the leading generalized eigenvector of $(\Ab,\Bb)$.
Let $V = \mathrm{supp}(\vb^*)$ be the index set corresponding to the non-zero elements of $\vb^*$, and let $|V| = s$.
Let $F \subset \{1,\ldots,d\}$ be a superset of $V$, i.e., $V\subset F$, with cardinality $|F|=k'$.  
Throughout the paper, for notational convenience, let $\lambda_j$ and $\hat{\lambda}_j$ be the $j$th generalized eigenvalue of the matrix pairs $(\Ab,\Bb)$ and $(\hat{\Ab},\hat{\Bb})$, respectively. 
Moreover, let $\lambda_j(F)$ and $\hat{\lambda}_j(F)$ be the $j$th generalized eigenvalue of the matrix pair $(\Ab_F,\Bb_F)$ and $(\hat{\Ab}_F,\hat{\Bb}_F)$, respectively. 

 Our theoretical results depend on several quantities that are specific to the  generalized eigenvalue problem.
Let 
\begin{equation}
\label{Eq:crawford}
\mathrm{cr}(\Ab,\Bb) = \underset{\vb: \|\vb\|_2=1}{\min} \; \left[(\vb^T\Ab \vb)^2+(\vb^T\Bb \vb)^2 \right]^{1/2}> 0
\end{equation} 
be the Crawford number of the symmetric-definite matrix pair $(\Ab,\Bb)$ \citep{stewart1979pertubation}.  Let
\begin{equation}
\label{Eq:inf crawford}
\mathrm{cr}(k') = \inf_{F : |F|\le k'} \mathrm{cr}(\Ab_F,\Bb_F)\qquad  \mathrm{and}\qquad 
\epsilon(k') = \sqrt{ \rho(\Eb_{\Ab},k')^2+\rho(\Eb_{\Bb},k')^2},
 \end{equation}
 where  $\rho(\Eb_{\Ab},k')$  is as defined in~(\ref{eq:w3}). 
   In the following, 
we start with an assumption that these quantities are upper bounded for sufficiently large $n$. 
\begin{assumption}
\label{ass:large n}
 For sufficiently large $n$, there exist constants $b,c >0$ such that 
\[
\frac{\epsilon(k')}{\mathrm{cr}(k')}\le b \qquad \mathrm{and}\qquad  \rho(\Eb_{\Bb},k') \le c \lambda_{\min} (\Bb)
\]
for any $k' \ll n$, where $\mathrm{cr}(k')$ and $\epsilon(k')$ are defined in \eqref{Eq:inf crawford}.
\end{assumption}
Provided that $n$ is large enough, it can be shown that the Assumption~\ref{ass:large n} holds~with high probability for most statistical models.  
In fact, we will show in Proposition~\ref{prop:concentration} in Section~\ref{subsec:application} that as long as $n > C  k' \log d$ for some sufficiently large constant $C$, then Assumption~\ref{ass:large n} is satisfied with high probability for most statistical models.
We will use the following implications of Assumption~\ref{ass:large n} in our theoretical analysis, which~are~implied by matrix perturbation theory \citep{stewart1979pertubation,stewart1990}. In detail, by applications of Lemmas~\ref{lemma:eigenvalue} and~\ref{lemma:perturbed pair} in Appendix \ref{proof:theorem:main}, we have that for any $F\subset \{1,\ldots,d\}$ with $|F|=k'$, there exist constants $a,c$ such that  
\[
(1-a) \lambda_j (F)\le \hat{\lambda}_j (F) \le (1+a)\lambda_j(F),
\qquad 
(1-c) \lambda_{j} (\Bb_F) \le \lambda_{j} (\hat{\Bb}_F) \le (1+c)\lambda_{j} (\Bb_F),
\]
and
\begin{equation}
\label{eq:kappa}
c_{\mathrm{lower}}\cdot \kappa(\Bb)
\le \kappa(\hat{\Bb}_F) \le c_{\mathrm{upper}} \cdot \kappa(\Bb),
\end{equation}
where $c_{\mathrm{lower}}=(1-c)/(1+c)$, $c_{\mathrm{upper}}= (1+c)/(1-c)$,  $c$ is the same constant  in Assumption \ref{ass:large n}, and $\kappa(\Bb)$ is the condition number of the matrix $\Bb$. Meanwhile, let $\gamma  = (1+a) \lambda_2/[(1-a) \lambda_1]$.

Finally, we define $\vb(F)$ to be the solution of a 
generalized eigenvalue problem restricted to a superset of $V$ ($V\subset F$):
\begin{equation}
\label{Eq:vbF}
\vb(F) = \underset{\vb \in \RR^d}{\arg \max} \;  \vb^T \hat{\Ab} \vb, \qquad \mathrm{subject \; to\; } \vb^T \hat{\Bb} \vb = 1, \quad \mathrm{supp}(\vb) \subseteq F. 
\end{equation}
The quantity $\vb(F)$ can be interpreted as the solution of a generalized eigenvalue problem for a low-dimensional problem when $k' < n$.
In the following theorem, we present our main theoretical result for Algorithm~\ref{alg:tgd} as a function of the $\ell_2$ distance between $\vb(F)$ and $\vb^*$.

\begin{theorem}
\label{theorem:main}
Let $k' = 2k+s$ and choose $k=Cs$ for sufficiently large $C$.  In addition, choose $\eta$ such that $\eta\lambda_{\max} (\Bb)<1/(1+c)$ and 
\[
\nu = \sqrt{1+2 [(s/k)^{1/2} + s/k] }\cdot \sqrt{
1-\frac{1+c}{8} \cdot \eta \cdot \lambda_{\min} ({\Bb})\cdot 
\left[ \frac{1-\gamma}{ c_{\mathrm{upper}} \kappa ({\Bb})+\gamma}\right]} < 1.
\] 
Input an initial vector $\vb_0$ with $\|\vb_0\|_2=1$ satisfying $|(\vb^*)^T\vb_0|/\|\vb^*\|_2 \ge 1- \theta(\Ab,\Bb)$, 
where $\theta(\Ab,\Bb)$ is a quantity given in Lemma~\ref{lemma:key} that depends on the matrix pair $(\Ab,\Bb)$.
Under Assumption~\ref{ass:large n}, we have 
\begin{equation}
\label{eq:w5} 
\sqrt{1- \frac{|(\vb^*)^T {\vb}_t|}{\|\vb^*\|_2}} \le \nu^t \cdot \sqrt{\theta(\Ab,\Bb)}+ \frac{\sqrt{20}}{1-\nu} \cdot \sqrt{1- \frac{|\vb(F)^T \vb^*|}{\|\vb(F)\|_2\|\vb^*\|_2}}.
\end{equation}
\end{theorem}

 For simplicity, assume that $(\vb^*)^T {\vb}_t$ is positive without loss of generality. Since ${\vb}_t$ is a unit vector, from \eqref{eq:w5} we have 
\[
1- \frac{|(\vb^*)^T {\vb}_t|}{\|\vb^*\|_2} = \frac{1}{2}\left\|\vb_t - \frac{\vb^*}{\|\vb^*\|_2} \right\|_2^2, \qquad 1- \frac{|\vb(F)^T {\vb}^*|}{\|\vb(F)\|_2 \|\vb^*\|_2} = \frac{1}{2}\left\|\frac{\vb(F)}{\|\vb(F)\|_2} - \frac{\vb^*}{\|\vb^*\|_2} \right\|_2^2.
\]
Thus, \eqref{eq:w5} states that the $\ell_2$ distance between $\vb^*/\|\vb^*\|_2$ and $\vb_t$ can be upper bounded by two terms. The first term on the right-hand side of \eqref{eq:w5} quantifies the optimization error, which decreases to zero at a geometric rate since $\nu<1$. Meanwhile, the second term on the right-hand side of \eqref{eq:w5} is the statistical error introduced for solving generalized eigenvalue problem restricted to the set $F$ as in \eqref{Eq:vbF}.
The result in Theorem~\ref{theorem:main} depends on the estimation error between $\vb(F)$ and $\vb^*$.
The following corollary quantifies such estimation error for a general class of symmetric-definite matrix pair $(\Ab,\Bb)$.


\begin{corollary}
\label{corollary1}
For a general class of symmetric-definite matrix pair~$(\Ab,\Bb)$, let
 \begin{equation}
 \label{Eq:eigengap}
 \Delta \lambda = \underset{j>1}{\min} \; \frac{\lambda_1-(1+a)\lambda_j}{\sqrt{1+\lambda_1^2} \sqrt{1+(1-a)^2 \lambda_j^2}}
 \end{equation}
 denote the eigengap for the generalized eigenvalue problem \citep{stewart1979pertubation,stewart1990}.
Assume that $\Delta \lambda>  \epsilon(k') /  \mathrm{cr}(k')$.  Then, under the same conditions as in Theorem~\ref{theorem:main}, we have 
\[
\sqrt{1- \frac{|(\vb^*)^T {\vb}_t|}{\|\vb^*\|_2}} \le \nu^t \cdot \sqrt{\theta(\Ab,\Bb)}+ \frac{\sqrt{10}}{1-\nu} \cdot 
\frac{2}{\Delta \lambda \cdot (\mathrm{cr}(k')-\epsilon(k'))}\cdot \epsilon(k'),
\]
where $\epsilon(k') = \sqrt{ \rho(\Eb_{\Ab},k')^2+\rho(\Eb_{\Bb},k')^2}$.
\end{corollary}
For a large class of statistical models, $\epsilon(k')$ converges to zero at the rate of $\sqrt{s\log d/n}$ with high probability.

\subsection{Theoretical Results for the Initialization in \eqref{eq:convexrelaxation}}
\label{subsec:initialization}
Theorem~\ref{theorem:main} involves a condition on the initialization $\vb_0$: the cosine angle between $\vb^*$ and $\vb_0$ needs to be strictly larger than a constant. 
In other words, the initialization $\vb_0$ needs to be close to $\vb^*$. 
We now present some theoretical guarantees for the initialization procedure in Section~\ref{convexrelax}.
In the context of sparse CCA, \citet{gao2014sparse} have shown that the estimated subspace obtained from solving convex relaxation of the form \eqref{eq:convexrelaxation} converges to the true subspace, under the assumption that $\Ab$ is low rank and positive semidefinite, and that the rank of $\Ab$ is known.
In the following proposition, we remove the aforementioned assumptions on $\Ab$.
Thus, a similar result holds more generally for the sparse generalized eigenvalue problem with symmetric-definite matrix pair $(\Ab,\Bb)$.

To this end, we define some additional notation.
Let $\Vb^*\in \RR^{d\times d}$ be $d$ generalized eigenvectors and let $\bLambda^* \in \RR^{d\times d}$ be a diagonal matrix of generalized eigenvalues of the matrix pair $(\Ab,\Bb)$, respectively.  
Let $\cS_v$ be a set containing indices of non-zero rows of $\Vb^*\in \RR^{d\times d}$.  For simplicity, assume that $|\cS_v| =s$ and that the eigenvalues of $\Bb$ are bounded.  
The matrix $\Ab$ can be rewritten in terms of its generalized eigenvectors and generalized eigenvalues up to sign jointly, $\Ab = \Bb \Vb^* \bLambda^* (\Vb^*)^T \Bb$ \citep{gao2014sparse}. 
Let $\tilde{\Ab} = \hat{\Bb} \Vb^* \bLambda^* (\Vb^*)^T \hat{\Bb}$ and let $\Pb^* = \Vb_{\cdot K}^*(\Vb_{\cdot K}^*)^T$, where $\Vb_{\cdot K}^*$ are the first $K$ generalized eigenvectors of $(\Ab,\Bb)$. Let $\hat{\Pb}$ be a solution to~\eqref{eq:convexrelaxation} with tuning parameters $\zeta$ and $K$.
The following proposition establishes an upper bound for the difference between $\hat{\Pb}$ and $\Pb^*$ under the Frobenius norm.

\begin{proposition}
\label{prop:initial}
Assume that $n$ is sufficiently large such that $\rho(\Eb_{\Bb},s^2) \le c \lambda_{\min} (\Bb)$, where $c$ is the same constant that appears in Assumption~\ref{ass:large n}.
Let  $\delta_{\mathrm{gap}} = \lambda_K-c\kappa(\Bb) \lambda_{K+1}/(1-c)$, and assume that $\delta_{\mathrm{gap}} > 0$.
Set  $\zeta > 2 \|\hat{\Ab}-\tilde{\Ab} \|_{\infty,\infty}$.
 Then,
 \[
 \|\hat{\Pb}-\Pb^*\|_F \le C\left(\frac{s}{\delta_{\mathrm{gap}} }\cdot \|\hat{\Ab}-\tilde{\Ab}\|_{\infty,\infty} + K\cdot \|\hat{\Bb}_{\cS_v}-\Bb_{\cS_v}\|_{2}  \right)  ,
 \] 
 where $C$ is a generic constant that does not depend on the generalized eigenvalues and the dimensions $n, d, s,$ and $K$.
\end{proposition}

 For most statistical models, it can be shown that $\|\hat{\Ab}-\tilde{\Ab} \|_{\infty,\infty}\le C_1 \sqrt{\log d/n}$ and $\|\hat{\Bb}_{\cS_v}-\Bb_{\cS_v}\|_{2} \le C_2 \sqrt{s/n}$ with high probability for generic constants $C_1$ and $C_2$. Thus, picking $\zeta > C_3 \sqrt{\log d/n}$,  the upper bound can be simplified to 
\[
\|\hat{\Pb}-\Pb^*\|_F \le C \left(\frac{s}{\delta_{\mathrm{gap}}  } \cdot\sqrt{\frac{\log d}{n}}+ K \sqrt{\frac{s}{n}}\right).
\]
 Choosing $K=1$ in \eqref{eq:convexrelaxation}, by a variant of the Davis-Kahan Theorem in \citet{vu2013fantope},
Proposition~\ref{prop:initial} guarantees that by setting $\vb_0$ to be 
the leading eigenvector of $\hat{\Pb}$, then $\vb_0$ will be sufficiently close to $\vb^*$ as long as the conditions in Proposition~\ref{prop:initial} are satisfied. In the next section, we will quantify the sample size condition needed for Proposition~\ref{prop:initial} to hold under various statistical models.

\subsection{Applications to Sparse PCA and Sparse CCA}
\label{subsec:application}
In this section, we provide some discussions on the implications of Theorem~\ref{theorem:main} and Proposition~\ref{prop:initial} in the context of sparse PCA and CCA, respectively.
More specifically, for each model, we first verify that the initial vector $\vb_0$ obtained from solving (\ref{eq:convexrelaxation}) is close to $\vb^*$. 
Therefore, the assumption on $\vb_0$ in Theorem~\ref{theorem:main} is satisfied. 
Next, we compare our results from Theorem~\ref{theorem:main} to the minimax optimal rate of convergence for each model.

\textbf{Sparse principal component analysis:}
We start with the sparse PCA problem.  We assume the model $\bX \sim N(\mathbf{0},\bSigma)$. 
As mentioned in Section~\ref{section:previous work}, sparse PCA is a special case of sparse generalized eigenvalue problem when $(\Ab,\Bb)= (\bSigma,\Ib)$ and $(\hat{\Ab},\hat{\Bb})=(\hat{\bSigma},\Ib)$, where $\hat{\bSigma}$ is the sample covariance matrix.
Thus, optimization problem \eqref{eq:convexrelaxation} reduces to a convex relaxation of sparse PCA proposed by \citet{vu2013fantope}. 
In this case, using a variant of the theoretical results in Proposition~\ref{prop:initial}, the initial value $\vb_0$ converges to $\vb^*$ as long as $n > C s^2 \log d$. 
Note that applying Corollary~\ref{corollary1} directly to the sparse PCA problem will give a loose upper bound (on the eigenfactor) since the additional information on the matrix pair $(\Ab,\Bb) = (\bSigma,\Ib)$, with $\Bb$ restricted to the identity matrix and  $\Ab$ restricted to positive definite matrix,  are not used in the derivation of Corollary~\ref{corollary1}. In other words, the results in Corollary~\ref{corollary1} are derived under a much larger class of matrix pair $(\Ab,\Bb)$.
To this end, we resort to the following corollary on the variant of Davis-Kahan perturbation result for sparse PCA (see, for instance, \citealp{yu2014useful}).

\begin{corollary}
\label{corollarypca}
Let $(\Ab,\Bb)=(\bSigma,\Ib)$ and let $\bSigma$ be a symmetric positive definite matrix. Let $\hat{\Ab}=\hat{\bSigma}$ be the sample covariance matrix.  We have 
\[
\rho(\hat{\Ab}-\Ab,s) \le  C \sqrt{\lambda_1(\Ab)} \sqrt{\frac{s \log d }{n}}
\]
holds with high probability for some constant $C>0$.
Suppose that $|F| = k'$ and that $k' = \cO(s)$.  Then, by the Davis-Kahan Theorem,
\[
\sqrt{1- \frac{|\vb(F)^T \vb^*|}{\|\vb(F)\|_2\|\vb^*\|_2}} \le C' \frac{\sqrt{\lambda_1(\Ab)}}{\lambda_1(\Ab)-\lambda_2(\Ab)} \sqrt{\frac{s\log d}{n}}
\]
holds with high probability for some constant $C'>0$. 
\end{corollary}

Combining Corollary~\ref{corollarypca} with Theorem~\ref{theorem:main}, our results indicate that as the optimization error decays to zero, our proposed estimator has a statistical rate of convergence of approximately 
\[
\frac{\sqrt{\lambda_1 (\Ab)}}{\lambda_1 (\Ab)-\lambda_2(\Ab)} \sqrt{\frac{s\log d}{n}},
\]
which matches the minimax optimal rate of convergence for sparse PCA problem \citep{cai2013sparse}.

\textbf{Sparse canonical correlation analysis:}
For sparse CCA, we assume the model:
\[
\begin{pmatrix}
\bX\\ \bY 
\end{pmatrix}\sim N (\mathbf{0},\bSigma) \qquad  \mathrm{and} \qquad
\bSigma = \begin{pmatrix}
\bSigma_x & \bSigma_{xy}\\ \bSigma_{xy}^T & \bSigma_y
\end{pmatrix}.
\] 
Recall from Example~\ref{example:cca} the definitions of $\hat{\Ab}$ and $\hat{\Bb}$ in the context of sparse CCA.   
The following proposition characterizes the rate of convergence between $\hat{\bSigma}$ and $\bSigma$. It follows from Lemma 6.5 of \citet{gao2014sparse}.  
Note that for the ease of presentation, we omit the dependence on the eigenvalues of $\Ab$ and $\Bb$ for CCA.

\begin{proposition}
\label{prop:concentration}
Let $\hat{\bSigma}_x$, $\hat{\bSigma}_y$, and $\hat{\bSigma}_{xy}$ be the sample  covariances of $\bSigma_x$, $\bSigma_y$, and $\bSigma_{xy}$, respectively. For any $C>0$ and positive integer $\overbar{k}$, there exists a constant $C'>0$ such that 
\[
\rho(\hat{\bSigma}_x - \bSigma_x ,\overbar{k}) \le  C\sqrt{\frac{\overbar{k}\log d}{n}},
\quad
\rho(\hat{\bSigma}_y - \bSigma_y ,\overbar{k}) \le  C\sqrt{\frac{\overbar{k}\log d}{n}},
\quad \mathrm{and} \quad
\rho(\hat{\bSigma}_{xy} - \bSigma_{xy},\overbar{k}) \le  C\sqrt{\frac{\overbar{k}\log d}{n}},
\]
with high probability. Moreover, $\|\hat{\bSigma}_{xy}- \bSigma_{xy}\|_{\infty,\infty} \le C\sqrt{\log d/n}$ with high probability.
\end{proposition}

We now verify the sample size condition in Proposition~\ref{prop:initial}.
From Proposition~\ref{prop:concentration}, we have $\rho(\Eb_{\Bb},s^2) = \cO_P(\sqrt{s^2\log d/n})$.  Thus, we need  $n > C s^2 \log d$ for some generic constant $C$.
Under the sample size condition and using the results in Proposition~\ref{prop:concentration}, it can be shown that $\|\tilde{\Ab}-\hat{\Ab}\|_{\infty,\infty} \le \|\tilde{\Ab}-{\Ab}\|_{\infty,\infty}+\|\hat{\Ab}-{\Ab}\|_{\infty,\infty} =   \cO_P(\sqrt{\log d /n})$.
Moreover, $\|\hat{\Bb}_{\cS_v}-\Bb_{\cS_v}\|_{2} = \cO_P(\sqrt{s /n})$.
Thus, as long as $n > C s^2 \log d$, 
 $\vb_0$ converges to $\vb^*$. 
This verifies the assumption on $\vb_0$ in Theorem~\ref{theorem:main}. 

In a recent paper by \citet{ma2016subspace}, the authors have shown that the minimax optimal eigenfactor takes the form $\sqrt{1-\lambda_1^2}\sqrt{1-\lambda_2^2}/(\lambda_1-\lambda_2)$ in the low-dimensional setting in which $n>d$, under the assumption that $\bSigma_x=\bSigma_y=\Ib$. Adapting the results in \citet{ma2016subspace} in a similar fashion as in Corollary~\ref{corollarypca}, Theorem~\ref{theorem:main} indicates that with high probability, our proposed estimator obtains the minimax statistical rate of convergence of approximately 
\begin{equation}
\label{minimax2}
\frac{\sqrt{1-\lambda_1^2}
\sqrt{1-\lambda_2^2}}{\lambda_1-\lambda_2}\cdot \sqrt{\frac{s \log d}{n}},
  \end{equation}
for the case when $\bSigma_x=\bSigma_y=\Ib$. However, the minimax optimal eigenfactor for  general $\bSigma_x$ and $\bSigma_y$ remains an open problem in the literature.

To obtain the rate of convergence for general $\bSigma_x$ and $\bSigma_y$, we will  apply Corollary~\ref{corollary1} to the sparse CCA problem.
Choosing $k$  to be of the same order as $s$, Proposition~\ref{prop:concentration} implies that both
 $\rho(\Eb_{\Ab}, k')$ and $\rho(\Eb_{\Bb},k')$ are at the order of $\sqrt{s\log d/n}$ with high probability. 
 Thus, Corollary~\ref{corollary1} indicates that as the optimization error decays to zero,  our proposed estimator has a  statistical rate of convergence of approximately
 \begin{equation}
 \label{minimax1}
\frac{\sqrt{1+\lambda_1^2}\sqrt{1+\lambda_2^2}}{\lambda_1-\lambda_2}   \cdot \sqrt{\frac{s \log d}{n}}. 
   \end{equation}
The upper bound is expected to be loose in terms of the eigenfactor since the class of paired matrices $(\Ab,\Bb)$ considered in Corollary~\ref{corollary1} is a much larger class of matrices than that of the sparse CCA.

In short, our theoretical results are very general and are not based on any statistical model. Moreover, the results in Theorem~\ref{theorem:main} are written as a function of the estimation error between $\vb(F)$, the solution of a generalized eigenvalue problem restricted on the set $F$, and $\vb^*$.  
Therefore, existing minimax optimal results for various statistical models in the low-dimensional setting can be adapted to the high-dimensional setting in a similar fashion as in the case of sparse CCA.

\section{Numerical Studies}
\label{section:simulation}
We perform extensive numerical studies to evaluate the performance of our proposal, Rifle, compared to existing methods.  
We consider 
sparse Fisher's discriminant analysis and sparse canonical correlation analysis, each of which can be recast as the sparse generalized eigenvalue problem (\ref{Eq:gep opt}), as shown in Examples~\ref{example:fda} and \ref{example:cca}.

Rifle involves an initial vector $\vb_0$ and a tuning parameter $k$ on the cardinality. 
We employ the convex optimization approach proposed in Section~\ref{convexrelax} to obtain an initial vector $\vb_0$. The convex approach involves two tuning parameters:  we simply select $\zeta = \sqrt{\log d /n}$ and $K=1$ as suggested by the theoretical analysis. Note that these tuning parameters can be selected conservatively since there is a refinement step to obtain a final estimator using Rifle.

It is challenging to propose a general model selection technique for the selection of $k$ in a sparse generalized eigenvalue problem since it is not based on any statistical model and it  includes both unsupervised learning and supervised learning methods as its special cases.  
 For supervised learning methods such as sparse FDA, we perform cross-validation to select the truncation parameter $k$.  For unsupervised learning methods such as the sparse PCA and CCA, it is generally agreed upon in the literature that model selection problem is challenging.  
In principle, we could also use cross-validation techniques to select $k$ in these settings such as the procedure considered in \citet{witten2009penalized}.  For simplicity, in our simulation studies, we assess the performance of our estimator in the context of sparse CCA across several values of $k$ and examine the role of $k$ under finite sample setting.

\subsection{Fisher's Discriminant Analysis}
\label{subsec:simFDA}
We consider high-dimensional classification problem using sparse Fisher's discriminant analysis.  
The data consists of an $n\times d$ matrix $\Xb$ with $d$ features measured on $n$ observations, each of which belongs to one of $K$ classes. 
We let $\xb_i$ denote the $i$th row of $\Xb$, and let $C_k \subset\{1,\ldots,n\}$ contains the indices of the observations in the $k$th class with $n_k = |C_k|$ and $\sum_{k=1}^K n_k = n$.

Recall from Example~\ref{example:fda} that this is a special case of the sparse generalized eigenvalue problem with $\hat{\Ab}= \hat{\bSigma}_b$ and $\hat{\Bb} = \hat{\bSigma}_w$. Let $\hat{\boldsymbol{\mu}}_k = \sum_{i\in C_k} \xb_i/n_k$ be the estimated mean for the $k$th class. The standard estimates for $\bSigma_w$ and $\bSigma_b$ are  
\[
\hat{\bSigma}_w =  \frac{1}{n} \sum_{k=1}^K \sum_{i\in C_k} (\xb_i-\hat{\boldsymbol{\mu}}_k)(\xb_i-\hat{\boldsymbol{\mu}}_k)^T \qquad \mathrm{and}\qquad 
\hat{\bSigma}_b = \frac{1}{n}\sum_{k=1}^K n_k \hat{\boldsymbol{\mu}}_k\hat{\boldsymbol{\mu}}_k^T.
\]
We consider two simulation settings similar to that of \citet{witten2009penalized}:
\begin{enumerate}
\item Binary classification: in this example, we set $\boldsymbol{\mu}_1 = \mathbf{0}$, $\mu_{2j} = 0.5$ for $j=\{2,4,\ldots,40\}$, and $\mu_{2j} =0$ otherwise.  Let $\bSigma$ be a block diagonal covariance matrix with five blocks, each of dimension $d/5\times d/5$. The $(j,j')$th element of each block takes value $0.8^{|j-j'|}$.  As suggested by \citet{witten2009penalized}, this covariance structure is intended to mimic the covariance structure of gene expression data.
The data are simulated as $\xb_i \sim N(\boldsymbol{\mu}_k,\bSigma)$ for $i\in C_k$.

\item Multi-class classification: there are $K=4$ four classes in this example.  Let $\mu_{kj} = (k-1)/3$ for $j=\{2,4,\ldots,40\}$ and $\mu_{kj} =0$ otherwise.  
The data are simulated as $\xb_i \sim N(\boldsymbol{\mu}_k,\bSigma)$ for $i\in C_k$, with the same covariance structure for binary classification.  As noted in \citet{witten2009penalized}, a one-dimensional vector projection of the data fully captures the class structure.
\end{enumerate}

Four approaches are compared: (i) Rifle; (ii) $\ell_1$-penalized logistic or multinomial regression implemented using the R package glmnet; (iii) $\ell_1$-penalized FDA with diagonal estimate of $\bSigma_w$ implemented using the R package penalizedLDA   \citep{witten2009penalized}; and (iv) direct approach to sparse  discriminant analysis \citep{mai2012direct,mai2015multiclass} implemented using the R package dsda and msda for binary and multi-class classification, respectively.  

For each method, models are fit on the training set with tuning parameter selected using 5-fold cross-validation.  Then, the models are evaluated on the test set. 
In addition to the aforementioned models, we consider an oracle estimator using the theoretical direction $\vb^*$, computed using the population quantities $\bSigma_w$ and $\bSigma_b$.

To compare the performance of the different proposals, we report the misclassification error on the test set and the number of non-zero features selected in the models.
The results for 400 training samples and 1000 test samples, with $d=500$ features, are reported in Table~\ref{Table:fda}.
From Table~\ref{Table:fda}, we see that Rifle has the lowest misclassification error compared to other competing methods.   
This suggests that Algorithm~\ref{alg:tgd} works well with the initial value obtained from the convex approach in Section~\ref{convexrelax}.
\citet{witten2009penalized} has the highest misclassification error in both of our simulation settings, since it does not take into account the dependencies among the features.   \citet{mai2012direct} and \citet{mai2015multiclass} perform slightly worse than our proposal in terms of misclassification error. Moreover, they use a large number of features in their model, which renders interpretation difficult.  
In contrast, the number of features selected by our proposal is very close to that of the oracle estimator.

\begin{table}[htp]
\begin{center}
\caption{The number of misclassified observations out of 1000 test samples and number of non-zero features (and standard errors) for binary and multi-class classification problems, averaged over 200 data sets. The results (rounded to the nearest integer) are for models trained with 400 training samples with 500 features.}
\begin{tabular}{cl   c ccc cc}
\hline
\hline
&&$\ell_1$-penalized& $\ell_1$-FDA&direct & Rifle&oracle \\ \hline
Binary  &Error & 32  (1)& 298 (1)  & 29 (1) & 15 (1)  &8 (1)     \\
&Features & 88 (1) & 23 (1) & 105 (2) &  42 (1)   & 41 (0)   \\ \hline
Multi-class  &Error & 495  (2)& 497 (1)  & 247 (2) & 192 (2)  &153 (1)     \\
&Features & 54 (2) & 22 (1) & 102 (2) &  42 (1)   & 41 (0)   \\ \hline
\end{tabular}
\label{Table:fda}
\end{center}
\end{table}

\subsection{Canonical Correlation Analysis}
\label{subsec:simCCA}
In this section, we study the relationship between two sets of random variables $\bX \in \RR^{d/2}$ and $\bY\in\RR^{d/2}$  in the high-dimensional setting using sparse CCA.  
Let $\bSigma_x$, $\bSigma_y$, and $\bSigma_{xy}$ be the covariance matrices of $\bX$ and $\bY$, and cross-covariance matrix of $\bX$ and $\bY$, respectively.  
We consider two different scenarios in which $\bSigma_{xy}$ is low rank and approximately low rank, respectively.

Throughout the simulation studies, we compare our proposal to \citet{witten2009penalized}, implemented using the R package PMA. Their proposal involves choosing two tuning parameters that controls the sparsity of the estimated directional vectors. We consider a range of tuning parameters and choose tuning parameters that yield the lowest estimation error for \citet{witten2009penalized}.  
 We assess the performance of Rifle by considering multiple values of $k=\{6,8,10,15\}$.  

The output of both our proposal and that of \citet{witten2009penalized} are normalized to have norm one, whereas the true parameters  $\vb_x^*$ and $\vb_y^*$ are normalized with respect to $\bSigma_x$ and $\bSigma_y$.
To evaluate the performance of the two methods, we normalize $\vb_x^*$ and $\vb_y^*$ such that they have norm one, and compute the squared $\ell_2$ distance between the estimated and the true directional vectors.    

\subsubsection{Low Rank $\mathbf{\Sigma}_{xy}$}

Assume that $(\bX,\bY) \sim N (\mathbf{0},\bSigma)$ with 
\[
\bSigma  = \begin{pmatrix} \bSigma_x & \bSigma_{xy} \\ \bSigma_{xy} & \bSigma_y \end{pmatrix} \qquad \mathrm{and} \qquad \bSigma_{xy} = \bSigma_x  \vb_x^* \lambda_1 (\vb_y^*)^T   \bSigma_y,
\]
where $0<\lambda_1<1$ is the largest generalized eigenvalue and $\vb_x^*$ and $\vb_y^*$ are the leading pair of canonical directions. 
The data consists of two $n\times ({d/2})$  matrices $\Xb$ and $\Yb$.
We assume that  each row of the two matrices are generated according to  $(\xb_i,\yb_i) \sim N(\mathbf{0},\bSigma)$.
The goal of CCA is to estimate the canonical directions $\vb^*_x$ and $\vb_y^*$ based on the data matrices $\Xb$ and $\Yb$.

Let $\hat{\bSigma}_x$, $\hat{\bSigma}_y$ be the sample covariance matrices of $\bX$ and $\bY$, and let $\hat{\bSigma}_{xy}$ be the sample cross-covariance matrix of $\bX$ and $\bY$. 
Recall from Example~\ref{example:cca} that the sparse CCA problem can be recast as the generalized eigenvalue problem with
\[
\hat{\Ab}= \begin{pmatrix}  \mathbf{0} & \hat{\bSigma}_{xy} \\ 
\hat{\bSigma}_{xy} & \mathbf{0}  \end{pmatrix},
\qquad 
\hat{\Bb}= \begin{pmatrix}  \hat{\bSigma}_{x} &\mathbf{0} \\ 
 \mathbf{0} & \hat{\bSigma}_{y}  \end{pmatrix},
\qquad \mathrm{and}\qquad
\vb = \begin{pmatrix} \vb_x \\ \vb_y
\end{pmatrix}.
\]
In our simulation setting, we set $\lambda_1 = 0.9$, $v_{x,j}^* = v_{y,j}^* =  1/\sqrt{3}$ for $j=\{1,6,11\}$, and $v_{x,j}^* = v_{y,j}^*  =0$ otherwise.  Then, we normalize $\vb_x^*$ and $\vb_y^*$ such that $(\vb_x^*)^T \bSigma_x \vb_x^* = (\vb_y^*)^T \bSigma_y \vb_y^*=1$.
We consider the case when $\bSigma_x$ and  $\bSigma_y$ are block diagonal matrix with five blocks, each of dimension $d/5\times d/5$, where the $(j,j')$th element of each block takes value $0.8^{|j-j'|}$.
The results for $d=500$, $s=6$, averaged over 200 data sets, are summarized in Table~\ref{table:CCA}.

\begin{table}[htp]
\begin{center}
\caption{Results for low rank $\bSigma_{xy}$. The squared $\ell_2$ distance between the estimated and true leading generalized eigenvector  as a function of the sample size $n$ for $d=500$, $s=6$. The results are averaged over 200 data sets.}
\begin{tabular}{cl   c ccc cc}
\hline
\hline
&&PMA& Rilfe ($k=6$)&Rilfe ($k=8$) & Rilfe ($k=10$)&Rilfe ($k=15$) \\ \hline
  &$n=200$ & 0.72  (0.01)& 0.21 (0.02)  & 0.11 (0.02) & 0.08 (0.02)  &0.07 (0.01) \\
$\vb_x$ &$n=400$ & 0.61 (0.01) & 0.01 (0.01) & 0.01 (0.01) &  0.01 (0.01)&0.01 (0.01)   \\ 
&$n=600$ &0.58 (0.01) & 0.01 (0.01) & 0.01 (0.01) &  0.01 (0.01)&0.01 (0.01)      \\ \hline
 &$n=200$ & 0.70  (0.01)& 0.24 (0.02)  & 0.24 (0.02) & 0.35 (0.02)  &0.58 (0.01)\\
$\vb_y$ &$n=400$ & 0.62 (0.01) & 0.02 (0.01) & 0.07 (0.01) &  0.15 (0.01)&0.32 (0.01)\\ 
 &$n=600$ & 0.59  (0.01)& 0.01 (0.01)  & 0.04 (0.01) & 0.08 (0.01)  &0.19 (0.01)\\\hline
\end{tabular}
\label{table:CCA}
\end{center}
\end{table}

From Table~\ref{table:CCA}, we see that our proposal outperforms \citet{witten2009penalized} uniformly across different sample sizes. This is not surprising since \citet{witten2009penalized} uses diagonal estimates of $\bSigma_x$ and $\bSigma_y$ to compute the directional vectors. 
The $\ell_2$ distance for our proposal decreases as we increase $n$.  Moreover, the $\ell_2$ distance increases when we increase $k$.  These results confirm our theoretical analysis in Theorem~\ref{theorem:main}.

\subsubsection{Approximately Low Rank $\mathbf{\Sigma}_{xy}$}
In this section, we consider the case when $\bSigma_{xy}$ is approximately low rank.  
We consider the same simulation set up as the previous section, except that $\bSigma_{xy}$ is now approximately low rank, generated as follows:
\[
 \bSigma_{xy} = \bSigma_x  \vb_x^* \lambda_1 (\vb_y^*)^T   \bSigma_y + \bSigma_{x} \Vb_x^* \bLambda (\Vb_y^*)^T \bSigma_y
\]
with $\lambda_1 = 0.9$.  Here, $\bLambda \in \RR^{200\times 200}$ is a diagonal matrix with diagonal entries equal 0.1, and $\Vb_x^*,\Vb_y^*\in \RR^{d/2\times 200}$ are normalized orthogonal matrices such that $(\Vb_x^*)^T \bSigma_x \Vb_x^* = \Ib$ and $( \Vb_y^*)^T\bSigma_y \Vb_y^*=\Ib$, respectively.  
The goal is to recover the leading generalized eigenvector $\vb_x^*$ and $\vb_y^*$.
The results for $d=1000, s=6$, averaged over 200 data sets, are summarized in Table~\ref{table:CCA2}.

From Table~\ref{table:CCA2}, we see that the performance for Rifle is much better than that of PMA across all settings.  As we increase the number of samples $n$, the $\ell_2$ distance decreases for all values of $k$.  Interesting, as we increase $k$ from $k=6$ to $k=10$ for the case when $n=400$, the $\ell_2$ distance decreases slightly.  
This is because in the high-dimensional setting, the initial value is not estimated accurately.  Thus, when we choose $k=s=6$, some of the true support are not selected after truncating the initial value $\vb_0$ and therefore it has a higher $\ell_2$ distance.  
In this case, by selecting a larger value of $k$, we are able to ensure that the true support are selected, which yields a lower $\ell_2$ distance.  
Note that if an even larger $k$ is selected, then the $\ell_2$ distance will eventually increase as in the case when $k=15$ for $\vb_y$.

\begin{table}[htp]
\begin{center}
\caption{Results for approximately low rank $\bSigma_{xy}$. The squared $\ell_2$ distance between the estimated and true leading generalized eigenvector  as a function of the sample size $n$ for $d=1000$, $s=6$. The results are averaged over 200 data sets.}
\begin{tabular}{cl   c ccc cc}
\hline
\hline
&&PMA& Rilfe ($k=6$)&Rilfe ($k=8$) & Rilfe ($k=10$)&Rilfe ($k=15$) \\ \hline
  &$n=400$ & 0.63  (0.01)& 0.30 (0.02)  & 0.19 (0.02) & 0.13 (0.02)  &0.07 (0.01) \\
$\vb_x$ &$n=600$ & 0.62 (0.01) & 0.11 (0.01) & 0.07 (0.01) &  0.09 (0.01)&0.07 (0.01)   \\ 
&$n=800$ &0.57 (0.01) & 0.02 (0.01) & 0.05 (0.01) &  0.08 (0.01)&0.07 (0.01)      \\ \hline
 &$n=400$ & 0.66  (0.01)& 0.31 (0.02)  & 0.26 (0.02) & 0.22 (0.02)  &0.25 (0.01)\\
$\vb_y$ &$n=600$ & 0.63 (0.01) & 0.10 (0.01) & 0.11 (0.01) &  0.13 (0.01)&0.16 (0.01)\\ 
 &$n=800$ & 0.55  (0.01)& 0.02 (0.01)  & 0.07 (0.01) & 0.11 (0.01)  &0.13 (0.01)\\\hline
\end{tabular}
\label{table:CCA2}
\end{center}
\end{table}

\section{Data Application}
In this section, we apply our method in the context of sparse sliced inverse regression as in Example~\ref{example:sir}. 
The data sets we consider are:
\begin{enumerate}
\item Leukemia \citep{golub1999molecular}: 7,129 gene expression measurements from 25 patients with acute myeloid leukemia and 47 patients with acute lymphoblastic luekemia. The data are available from {\tt{http://www.broadinstitute.org/cgi-bin/cancer/datasets.cgi}}. Recently, this data set is analyzed in the context of sparse sufficient dimension reduction in \citet{yin2015sequential}.  

\item Lung cancer \citep{spira2007airway}: 22,283 gene expression measurements from large airway epithelial cells sampled from 97 smokers with lung cancer and 90 smokers without lung cancer.  The data are publicly available from GEO at accession number GDS2771.  
\end{enumerate}

We preprocess the leukemia data set following \citet{golub1999molecular} and \citet{yin2015sequential}.  In particular, we set gene expression readings of 100 or fewer to 100, and expression readings of 16,000 or more to 16,000.  We then remove genes with difference and ratio between the maximum and minimum readings that are less than 500 and 5, respectively.  A log-transformation is then applied to the data.  This gives us a data matrix $\Xb$ with 72 rows/samples and 3571 columns/genes. For the lung cancer data, we simply select the 2,000 genes with the largest variance as in \citet{petersen2015fused}.
This gives a data matrix with 167 rows/samples and 2,000 columns/genes.  We further standardize both the data sets so that the genes have mean equals zero and variance equals one. 

\begin{figure}[!t]
\centering
\subfigure[]{\includegraphics[scale=0.55]{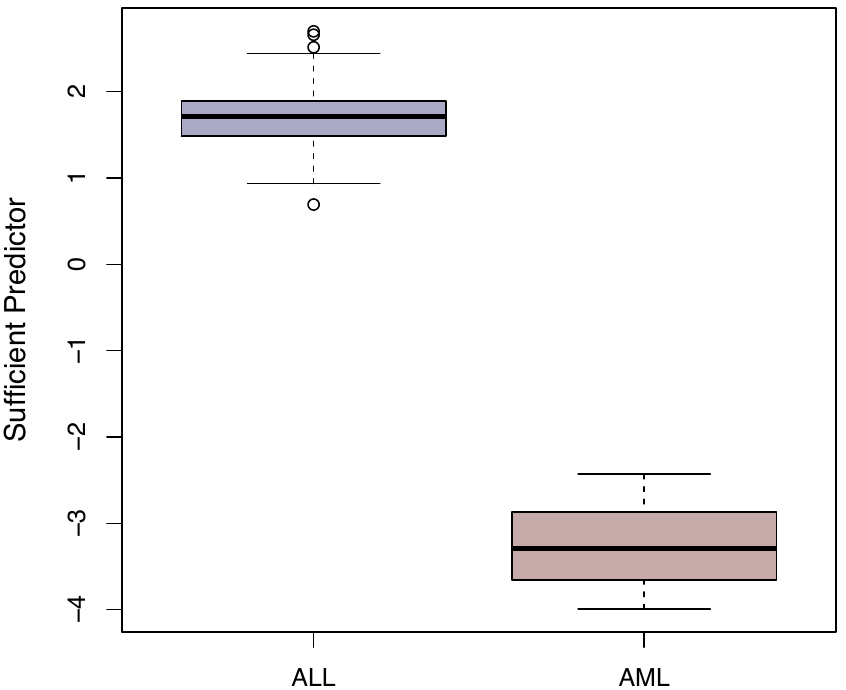}}
\subfigure[]{\includegraphics[scale=0.55]{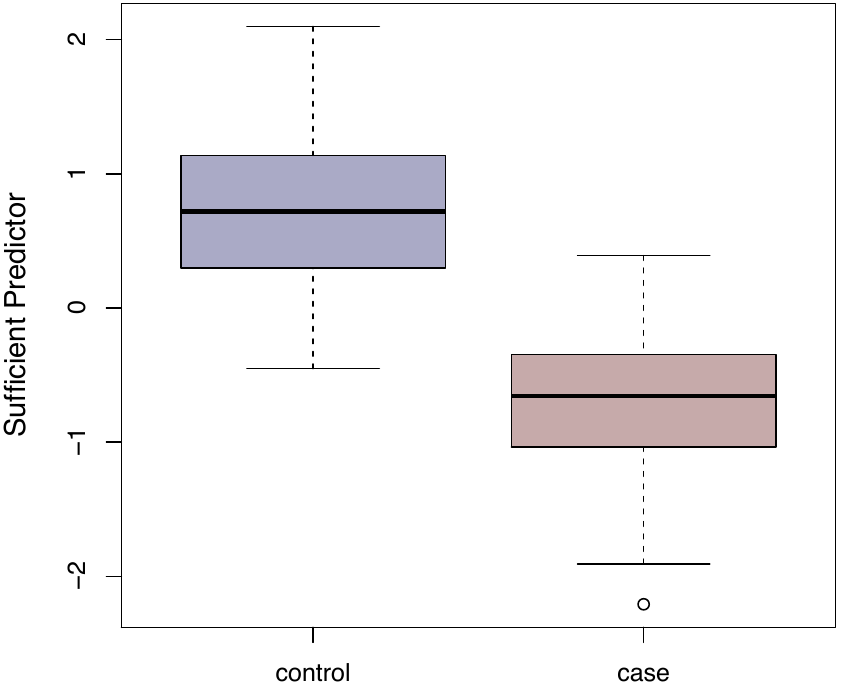}}
          \caption{Panels (a) and (b) contain box-plots of the sufficient predictor $\Xb \hat{\vb}_t$ obtained from Algorithm~\ref{alg:tgd} for the leukemia and lung cancer data sets. In panel (a), the $y$-axis represents patients with acute lymphoblastic leukemia (ALL) and  acute myeloid leukemia (AML), respectively.  In panel (b), the $y$-axis represents patients with and without lung cancer, respectively.             }
          \label{Fig:realdata}
\end{figure}

Recall from Example~\ref{example:sir} that in order to apply our method, we need the estimates $\hat{\Ab}= \hat{\bSigma}_{E(\bX\mid Y)}$ and $\hat{\Bb} = \hat{\bSigma}_{x}$.  The quantity $\hat{\bSigma}_x$ is simply the sample covariance matrix of $\bX$.
Let $n_1$ and $n_2$ be the number of samples of the two classes in the data set.  Let $\hat{\bSigma}_{x,1}$ and $\hat{\bSigma}_{x,2}$ be the sample covariance matrix calculated using only data from class one and class two, respectively.  
Then, the covariance matrix of the conditional expectation can be estimated by 
\[
\hat{\bSigma}_{E[\bX\mid Y]} = \hat{\bSigma}_{x}  - \frac{1}{n}\sum_{k=1}^2 n_k \hat{\bSigma}_{x,k},
\]
where $n=n_1+n_2$ \citep{li1991sliced,li2006sparse,zhu2006sliced,li2008sliced,chen2010coordinate,yin2015sequential}. 
Let $\hat{\vb}_t$ be the output of Algorithm~\ref{alg:tgd}.  Similar to \citet{yin2015sequential}, we plot the box-plot of the sufficient predictor, $\Xb \hat{\vb}_t$, for the two classes in each data set.  The results with $k=25$ for leukemia and lung cancer data sets are in Figures~\ref{Fig:realdata}(a)-(b), respectively.

From Figure~\ref{Fig:realdata}(a), for the leukemia data set, we see that the sufficient predictor for the two groups are much more well separated than the results in \citet{yin2015sequential}.  Moreover, our proposal is with theoretical guarantees whereas their proposal is sequential without theoretical guarantees.
For the lung cancer data set, we see that there is some overlap between the sufficient predictor for subjects with and without lung cancer.  These results are consistent in the literature where it is known that the lung cancer data set is a much more difficult classification problem compared to that of the leukemia data set \citep{fan2008high,petersen2015fused}.

\section{Discussion}
We propose a two-stage computational framework for solving the sparse generalized eigenvalue problem.
 The proposed method successfully handles ill-conditioned normalization matrix that arises from the high-dimensional setting due to finite sample estimation, and the final estimator enjoys geometric convergence to a solution with the optimal statistical rate of convergence. 
Our method and theory have applications to a large class of  statistical models including but are not limited to sparse FDA, sparse CCA, and sparse SDR. 
Compared to existing theory for each specific statistical model, our theory is very general and does not require any structural assumption on ($\Ab,\Bb)$.

Our theoretical results in Theorem~\ref{theorem:main} rely on selecting the tuning parameter $k$ such that $k = Cs$ for some constant $C>1$.  
However, in practice, the true sparsity level $s$ is unknown and it may be difficult to select the value of $k$.  To remove the dependencies on $s$, one of the reviewers suggested a thresholding strategy, i.e., instead of truncating the vector $\vb_t'$ and keeping the top $k$ elements, one can perform a $C \cdot \sqrt{\log d/n}$ thresholding on the updated vector $\vb_t'$ from Step 3 of Algorithm~\ref{alg:tgd}, where $C$ is some user-specified constant.  To evaluate the thresholding strategy, we perform a small scale numerical study on the FDA binary classification example similar to that of Section~\ref{subsec:simFDA} with $n = 200$ and $d=200$.  We compare the estimator obtained using the soft-thresholding rule (Soft-Rifle) and that of our proposed truncation rule by calculating the estimation error  between these estimators and the oracle direction. The results, averaged across 50 iterations, are presented in the Table~\ref{Table:fdasmall}. From Table~\ref{Table:fdasmall}, we see that depending on the choice of the constant $C$, the soft-thresholding rule have similar performance as the truncation rule, suggesting that substituting the soft-thresholding rule onto Steps 4 and 5 of Algorithm~\ref{alg:tgd} will also work. 

\begin{table}[htp]
\begin{center}
\caption{Estimation error between the true standardized generalized eigenvector ($\|\vb^* \|_2=1$) and the estimated generalized eigenvector for binary classification problem, averaged over 50 data sets. The  number of non-zero features are also reported.  The results are with $n=200$ and $d=200$. The true sparsity level is $s=40$.}
\vspace{1mm}
\begin{tabular}{   c| ccc|ccc|}
\hline
\hline
& & Soft-Rifle & & &Rifle &\\ 
&$C = 1$& $C=0.5$&$C=0.25$&$k=35$& $k=40$&$k=55$  \\ \hline
  Estimation Error & 0.180 & 0.048 & 0.072  & 0.181& 0.048& 0.072    \\ 
    Features & 33.5 & 39.7  & 53.3 &35&40&55     \\ \hline
\end{tabular}
\label{Table:fdasmall}
\end{center}
\end{table}

In the case when $\vb^*$ is approximately sparse, i.e., $s=d$, the current theoretical results are no longer applicable.  To address this issue, we can redefine the notion of sparsity level $s$.
As suggested by one of the reviewers, we can define the  effective sparsity level $s'$ as the $\ell_q$ norm $(q < 1)$ or the ratio between, for example, $\ell_1$ and $\ell_\infty$ norms of $\vb^*$. The theoretical properties for thresholding strategy and weak sparsity are challenging to establish under our current theoretical framework.  In particular, due to the normalization constraint $\vb^T \hat{\Bb} \vb$ on the denominator, to analyze the gradient ascent step in Step 2, we require that the cardinality of the input vector must have support $k'$.  
This condition is needed to control the condition number of $\hat{\Bb}_F$, where $F$ is an index set such that $|F| = k'$.    Developing a new theoretical framework for solving the sparse generalized eigenvalue problem is out of the scope of this paper and we leave it for future work.

There are several additional future directions for the sparse generalized eigenvalue problem. It will be interesting to study whether Rifle can be generalized to the case for estimating  subspace spanned by the top $K$ leading generalized eigenvectors.  
The computational bottleneck for the current approach is on the convex relaxation method for obtaining the initial vector $\vb_0$, which has a computational complexity of $\cO(d^3)$ per iteration. This yields a total computational complexity of $\cO(d^3) + \cO(kd+d)$ for the proposed two-stage computational framework.  In future work, it will be of paramount importance to propose an efficient convex algorithm to obtain $\vb_0$ such that our proposal is scalable to accommodate large-scale data.

\section*{Acknowledgement}
We thank the editor, associate editor, and two reviewers for their helpful comments that improve earlier version of this paper. We thank Gao Chao and Xiaodong Li for responding to our inquiries.   Tong Zhang was supported by NSF IIS-1250985, NSF IIS-1407939, and NIH R01AI116744. Kean Ming Tan was supported by  NSF IIS-1250985, NSF IIS-1407939, and NSF DMS-1811315.

\bibliographystyle{ims}
\bibliography{reference}

\newpage
\appendix
\section{Proof of Theorem~\ref{theorem:main}}
\label{proof:theorem:main}
To establish Theorem~\ref{theorem:main}, we first quantify the error introduced by maximizing the empirical version of the
generalized eigenvalue problem, restricted to a superset of $V$ ($V\subset F$), that is, 
\begin{equation*}
\vb(F) = \underset{\vb \in \RR^d}{\arg \max} \;  \vb^T \hat{\Ab} \vb, \qquad \mathrm{subject \; to\; } \vb^T \hat{\Bb} \vb = 1, \quad \mathrm{supp}(\vb) \subseteq F. 
\end{equation*}
Then we establish an error bound between $\vb_t'$ in Step 2 of Algorithm~\ref{alg:tgd} and $\vb(F)$. Finally, we quantify the error introduced by the truncated step in Algorithm~\ref{alg:tgd}.

We first state a series of lemmas that will facilitate the proof of Theorem~\ref{theorem:main}.  The proofs for the technical lemmas are deferred  to Appendix~\ref{appendixB}.
We start with some results from perturbation theory for eigenvalue and generalized eigenvalue problems \citep{golub2012matrix}.

\begin{lemma}
\label{lemma:eigenvalue}
Let $\Jb$ and $\Jb+\Eb_{\Jb}$ be $d\times d$ symmetric matrices. Then, for all $k\in \{1,\ldots,d\}$, 
\[
 \lambda_k(\Jb)+\lambda_{\min}(\Eb_{\Jb}) \le \lambda_k (\Jb+\Eb_{\Jb}) \le \lambda_k(\Jb)+\lambda_{\max}(\Eb_{\Jb}).
\] 
\end{lemma}
\noindent
In the sequel, we state a result on the perturbed generalized eigenvalues for a symmetric-definite matrix pair $(\Jb,\Kb)$  
in the following lemma, which follows directly from Theorem 3.2 in \citet{stewart1979pertubation} and Theorem 8.7.3 in \citet{golub2012matrix}.  
\begin{lemma}
\label{lemma:perturbed pair}
Let $(\Jb,\Kb)$ be a symmetric-definite matrix pair with generalized eigenvalues $\lambda_1( \Jb,  \Kb) \ge \cdots \ge \lambda_d( \Jb,  \Kb) $.
Let ($ \Jb+\Eb_{\Jb}$,$ \Kb+\Eb_{\Kb}$) be the perturbed matrix pair and assume that $\Eb_{\Jb}$ and $\Eb_{\Kb}$ satisfy  
\[
\epsilon =\sqrt{ \|\Eb_{\Jb}\|_{2}^2 + \|\Eb_{\Kb}\|_{2}^2}   < \mathrm{cr}(\Jb,\Kb),
\]
where $\mathrm{cr}(\Jb,\Kb)$ is as defined in (\ref{Eq:crawford}).  
Then, $( \Jb+\Eb_{\Jb}$, $ \Kb+\Eb_{\Kb})$ is a symmetric-definite matrix pair with generalized eigenvalues $\lambda_1( \Jb+\Eb_{\Jb},  \Kb+\Eb_{\Kb}) \ge \cdots \ge \lambda_d( \Jb+\Eb_{\Jb},  \Kb+\Eb_{\Kb}) $.  
  Then,
\[
\frac{ \lambda_k( \Jb,  \Kb)  \cdot \mathrm{cr}(\Jb,\Kb)-\epsilon}{\mathrm{cr}(\Jb,\Kb) + \epsilon\cdot  \lambda_k( \Jb,  \Kb) } \le {\lambda}_k (  \Jb+\Eb_{\Jb},  \Kb+\Eb_{\Kb})  \le \frac{\lambda_k( \Jb,  \Kb) \cdot \mathrm{cr}(\Jb,\Kb)+\epsilon}{\mathrm{cr}(\Jb,\Kb) - \epsilon\cdot  \lambda_k( \Jb,  \Kb)  }.
\]
\end{lemma}
Recall from Section~\ref{section:theory} that $\vb^*$ is the first generalized eigenvector of $(\Ab,\Bb)$ with generalized eigenvalue $\lambda_1$, and that $V=\mathrm{supp}(\vb^*)$.   
For any given set $F$ such that $V\subset F$, let ${\lambda}_k(F)$  and $\hat{\lambda}_k(F)$ be the $k$th generalized eigenvalues of  $(\Ab_F,\Bb_F)$  and $(\hat{\Ab}_F,\hat{\Bb}_F)$, respectively.
Under Assumption~\ref{ass:large n} and by an application Lemma~\ref{lemma:perturbed pair}, we have
\[
\frac{\hat{\lambda}_2(F)}{\hat{\lambda}_1(F)} \le \gamma,
\] 
 where $\gamma = (1+a)\lambda_2 /[(1-a)\lambda_1]$.

Let $\yb(F) = \vb(F)/\|\vb(F)\|_2$ and~$\yb^* = \vb^*/\|\vb^*\|_2$ such that $\|\yb(F)\|_2 = \|\yb^*\|_2=1$. 
We now present a key lemma on measuring the progress of the gradient descent step.  It requires an initial solution that is close enough to the optimal value  in (\ref{Eq:vbF}). With some abuse of notation, we indicate $\yb(F)$ to be a $k'$-dimensional vector restricted to the set $F \subset \{1,\ldots,d\}$ with $|F| = k'$. Recall that $c>0$ is some arbitrary small constant stated in Assumption~\ref{ass:large n} and $c_{\mathrm{upper}}$ is defined as $(1+c)/(1-c)$.

\begin{lemma}
\label{lemma:key}
Let $F\subset \{1,\ldots,d\}$ be some set with $|F| = k' $.  
  Given any $\tilde{\vb}$ such that $\|\tilde{\vb}\|_2 =1$ and $\tilde{\vb}^T \yb(F) >0$, 
let $\rho = \tilde{\vb}^T \hat{\Ab}_F\tilde{\vb}/ \tilde{\vb}^T{\hat{\Bb}_F} \tilde{\vb}$, and let $\vb' = \Cb_F\tilde{\vb} /\|\Cb_F \tilde{\vb}\|_2$, where 
\$
\Cb= \Ib + (\eta /\rho) (\hat{\Ab}-\rho \hat{\Bb})
\$ and $\eta>0$ is some positive constant.  
Let $\delta =1-\yb(F)^T \tilde{\vb}$.  
Pick $\eta$ sufficiently small such that 
  \$
  \eta \lambda_{\max} ({\Bb})  < 1/(1+c),
  \$ and $\delta$ is sufficiently small such that 
 \[
1-\delta \ge1- \theta(\Ab,\Bb),
 \]
 where  
 \[
\theta(\Ab,\Bb)= \min \left(  \frac{1}{8 c_{\mathrm{upper}}\kappa(\Bb)}, \frac{1/\gamma-1}{3c_{\mathrm{upper}}\kappa(\Bb)}   ,
  \frac{1-\gamma}{30\cdot (1+c) \cdot c_{\mathrm{upper}}^2\cdot \eta \cdot \lambda_{\max} (\Bb) \cdot \kappa^2(\Bb) \cdot [c_{\mathrm{upper}}\kappa(\Bb)+\gamma]}
  \right).
  \]
 Then under Assumption~\ref{ass:large n}, we have 
\[
\yb(F)^T \vb' \ge\yb(F)^T \tilde{\vb} +\frac{1+c}{8} \cdot  \eta\cdot  \lambda_{\min} (\Bb)\cdot  [1-\yb(F)^T \tilde{\vb}]\cdot  \left[ \frac{1-\gamma}{ c_{\mathrm{upper}} \kappa ({\Bb})+\gamma}\right].
\]
\end{lemma}

The following lemma characterizes the error introduced by the truncation step. It follows directly from Lemma 12 in \citet{yuan2013truncated}.

\begin{lemma}
\label{lemma:truncation}
Consider $\yb'$ with $F'=\mathrm{supp}(\yb')$ and $ |F'| = \overbar{k}$. Let $F$ be the indices of ${\yb}$ with~the largest $k$ absolute values, with $|F|=k$. If $\|\yb'\|_2 = \|{\yb}\|_2 = 1$, then 
\$
&|\mathrm{Truncate}(\yb,F)^T {\yb}'| \\
&\quad\ge |\yb^T \yb'| - (\overbar{k}/k)^{1/2} \min \left( 
\sqrt{1-(\yb^T \yb')^2}, [1+(\overbar{k}/k)^{1/2}]\cdot [1-(\yb^T \yb')^2]
\right).
\$

\end{lemma}

Recall from Algorithm~\ref{alg:tgd} that we define $\vb_t = \hat{\vb}_t /\|\hat{\vb}_t\|_2$.  
Since $\|\vb_t'\|_2=1$, and $\hat{\vb}_t$ is the truncated version of $\vb_t'$, we have that $\|\hat{\vb}_t\|_2\leq 1$. This implies that $|(\yb^*)^T \vb_t|  \ge  |(\yb^*)^T \hat{\vb}_t|$.  We now quantifies the progress of each iteration of Algorithm~\ref{alg:tgd}.  
To this end, assume that $k> s$, where $s$ is the cardinality of the support of $\yb^*=\vb^*/\|\vb^*\|_2^2$, and $k$ is the truncation parameter in Algorithm~\ref{alg:tgd}.  Let $k' = 2k + s$ and let 
\[
\nu = \sqrt{1+2 [(s/k)^{1/2} + s/k] }\cdot \sqrt{
1-\frac{1+c}{8} \cdot\eta \cdot \lambda_{\min} ({\Bb})\cdot 
\left[ \frac{1-\gamma}{c_{\mathrm{upper}}\kappa ({\Bb})+\gamma}\right]}.
\]

Recall that $V$ is the support of $\vb^*$, the population leading generalized vector and also $\yb^* = \vb^*/\|\vb^*\|_2$.    Let $F_{t-1} = \mathrm{supp}(\vb_{t-1})$, $F_{t} = \mathrm{supp}(\vb_{t})$,
and let   $F=F_{t-1}\cup F_t \cup V$. Note that the cardinality of $F$ is no more than $k' = 2k+s$, since $|F_t|=|F_{t-1}| = k$.  Let 
\[
\vb_t' = \Cb_F \vb_{t-1}/ \| \Cb_F \vb_{t-1}\|_2,
\]
where $\Cb_F$ is the submatrix of $\Cb_F$ restricted to the rows and columns indexed by $F$.
We note that ${\vb}_t'$ is equivalent to the one in Algorithm~\ref{alg:tgd}, since the elements of $\vb_t'$ outside of the set $F$ take value zero.
Without loss of generality and for simplicity, we assume that the inner product between two eigenvectors are positive, because otherwise we can simply do appropriate sign changes in the proof.

Applying Lemma~\ref{lemma:key} with the set $F$, we obtain 
\[
\yb(F)^T \vb_t' \ge \yb(F)^T \vb_{t-1} + \frac{1+c}{8} \cdot  \eta\cdot  \lambda_{\min} ({\Bb})\cdot  [1-\yb(F)^T {\vb_{t-1}}] \cdot  \left[ \frac{1-\gamma}{c_\mathrm{upper}\kappa ({\Bb})+\gamma}\right].
\]
Subtracting both sides of the equation by one and rearranging the terms, we obtain 
\begin{equation}
\label{Eq:lemma:combine}
1-\yb(F)^T \vb_t'  \le [ 1-  \yb(F)^T \vb_{t-1} ] \cdot \left\{ 
1-\frac{1+c}{8}\cdot \eta \cdot \lambda_{\min} ({\Bb})\cdot 
\left[ \frac{1-\gamma}{c_\mathrm{upper}\kappa ({\Bb})+\gamma}\right]\right\}.
\end{equation}
This implies that
\begin{equation}
\label{Eq:lemma:combine2}
\|\yb(F)- \vb_t'\|_2  \le \|  \yb(F)- \vb_{t-1} \|_2 \cdot \sqrt{
1-\frac{1+c}{8}\cdot \eta \cdot \lambda_{\min} ({\Bb})\cdot 
\left[ \frac{1-\gamma}{c_\mathrm{upper}\kappa ({\Bb})+\gamma}\right]}.
\end{equation}

By the triangle inequality, we have 
\begin{equation}
\label{Eq:lemma:combine3}
\begin{split}
\|\yb-\vb_t'\|_2 &\le \|\yb(F)- \vb_t'\|_2+ \|\yb(F)- \yb^*\|_2\\
&\le \|  \yb(F)- \vb_{t-1} \|_2 \cdot \sqrt{
1-\frac{1+c}{8}\cdot \eta \cdot \lambda_{\min} ({\Bb})\cdot 
\left[ \frac{1-\gamma}{c_\mathrm{upper}\kappa ({\Bb})+\gamma}\right]}+ \|\yb(F)- \yb^*\|_2\\
&\le \|  \yb- \vb_{t-1} \|_2 \cdot \sqrt{
1-\frac{1+c}{8}\cdot \eta \cdot \lambda_{\min} ({\Bb})\cdot 
\left[ \frac{1-\gamma}{c_\mathrm{upper}\kappa ({\Bb})+\gamma}\right]}+2\|\yb(F)- \yb^*\|_2,
\end{split}
\end{equation}
where the second inequality follows from~(\ref{Eq:lemma:combine2}).  This is equivalent to 
\begin{equation}
\label{Eq:lemma:combine3}
\sqrt{1- |\yb^T \vb_t'| } \le \sqrt{1- |\yb^T \vb_{t-1}| }   \cdot \sqrt{
1-\frac{1+c}{8} \cdot \eta \cdot \lambda_{\min} ({\Bb})\cdot 
\left[ \frac{1-\gamma}{c_\mathrm{upper}\kappa ({\Bb})+\gamma}\right]}+2\cdot \sqrt{1- |\yb(F)^T \yb^*|}.
\end{equation}

We define 
\[
\nu = \sqrt{1+2 [(s/k)^{1/2} + s/k] }\cdot \sqrt{
1-\frac{1+c}{8} \cdot\eta \cdot \lambda_{\min} ({\Bb})\cdot 
\left[ \frac{1-\gamma}{c_\mathrm{upper}\kappa ({\Bb})+\gamma}\right]}.
\] 
By Lemma~\ref{lemma:truncation} and picking $k>s$, we have 
\begin{equation}
\label{Eq:lemma:combine4}
\begin{split}
\sqrt{1- |\yb^T \hat{\vb}_t| }  &\le  \sqrt{1- |\yb^T \vb_t'| +
[(s/k)^{1/2} + s/k] \cdot [1-|\yb^T \vb_t'|^2]    }\\
&\le \sqrt{1- |\yb^T \vb_t'|} \cdot \sqrt{1+ [(s/k)^{1/2} + s/k] \cdot [1+|\yb^T \vb_t'|]}\\
&\le \sqrt{1- |\yb^T \vb_t'|} \cdot \sqrt{1+2 [(s/k)^{1/2} + s/k] }\\
&\le  \nu \sqrt{1- |\yb^T \vb_{t-1}|} + \sqrt{20} \cdot \sqrt{1- |\yb(F)^T \yb^*|},
\end{split}
\end{equation}
where the third inequality holds using the fact that $|\yb^T \vb'_t|\le 1$, and the last inequality holds by (\ref{Eq:lemma:combine3}).

Finally, we have 
\begin{equation}
\label{Eq:proof theorem main1}
\begin{split}
\sqrt{1- |(\yb^*)^T {\vb}_t| }&\le \sqrt{1- |(\yb^*)^T \hat{\vb}_t| }\\
&\le \nu \sqrt{1- |(\yb^*)^T \vb_{t-1}|} + \sqrt{20} \cdot \sqrt{1- |\yb(F)^T \yb^*|}.
\end{split}
\end{equation}
By recursively applying \eqref{Eq:lemma:combine4}, we have for all $t\ge 0$, 
\[
\sqrt{1- |(\yb^*)^T {\vb}_t|} \le  \nu^t \sqrt{1- |(\yb^*)^T \vb_{0}|} +\sqrt{20} \cdot \sqrt{1- |\yb(F)^T \yb^*|}/(1-\nu),
\]
as desired.

\section{Proof of Corollary 1}
\label{appendixC}
Let $F\supset V$ be a superset of the support of $\yb^*$.  Recall that $\yb(F) = \vb(F)/\|\vb(F)\|_2$ and $\yb^* = \yb^*/\|\yb^*\|_2$. We first prove that $\yb(F)$ is close to $\yb^*$ for a general class of symmetric-definite matrix pair $(\Ab,\Bb)$. 
To this end, we present the following lemma resulting from Theorem 4.3 in \citet{stewart1979pertubation}.

\begin{lemma}
\label{lemma:perturbation vec}
Let $F$ be a set such that $ V\subset F$ with $|F|=k'>s$ and let  
\[
 \delta (F) = \sqrt{\|\Eb_{\Ab,F}\|_2^2+\|\Eb_{\Bb,F}\|_2^2 }.
\]
 Let 
\[
\chi(\lambda_1(F),\hat{\lambda}_k(F)) = \frac{|\lambda_1(F)-\hat{\lambda}_k(F)|}{\sqrt{1+\lambda_1(F)^2}\cdot \sqrt{1+\hat{\lambda}_k(F)^2}};  \qquad \Delta \hat{\lambda}(F) = \underset{k> 1}{\min} \;\chi (\lambda_1(F),\hat{\lambda}_k(F)) >0.
\]
If $\delta(F)/\Delta \hat{\lambda}(F) < \mathrm{cr}(\hat{\Ab}_F,\hat{\Bb}_F)$, then 
\[
\frac{\min \{\|\vb(F) -\vb^*\|_2 ,\vb(F) +\vb^*\|_2  \}}{\|\vb^*\|_2}\le \frac{\delta(F)}{ \Delta \hat{\lambda} (F)\cdot \mathrm{cr}(\hat{\Ab}_F,\hat{\Bb}_F) }.
\]
This implies that  
\[
\min \{\|\yb(F) -\yb^*\|_2,\|\yb(F) +\yb^*\|_2\}\le \frac{2 }{ \Delta \lambda \cdot (\mathrm{cr}(k')-\epsilon(k'))}\cdot \epsilon(k'),
\]
where $\Delta \lambda$, $\mathrm{cr}(k')$, and $\epsilon(k')$ are as defined in (\ref{Eq:eigengap}) and (\ref{Eq:inf crawford}).
\end{lemma}
By Lemma~\ref{lemma:perturbation vec}, we have 
\[
\sqrt{1- \frac{|(\vb^*)^T {\vb}_t|}{\|\vb^*\|_2}} \le \frac{2^{1/2} }{ \Delta \lambda \cdot (\mathrm{cr}(k')-\epsilon(k'))}\cdot \epsilon(k').
\]
Substituting the above inequality into Theorem~\ref{theorem:main} yields the results in Corollary~\ref{corollary1}.

\section{Proof of Technical Lemmas}
\label{appendixB}

\subsection{Proof of Lemma~\ref{lemma:key}}
\begin{proof}
Recall that $F\subset \{1,\ldots,d\}$ is some set with cardinality $|F| = k'$.  Also, recall that  $\yb(F)$  is proportional to the largest generalized eigenvector of $(\hat{\Ab}_F,\hat{\Bb}_F)$.
Throughout the proof,  we write $\hat{\kappa}$ to denote $\kappa (\hat{\Bb}_F)$ for notational convenience. In addition, we use the notation $\|\vb\|_{\hat{\Bb}_F}^2$ to indicate $\vb^T \hat{\Bb}_F \vb$.

Let $\bxi_j$ be the $j$th generalized eigenvector of $(\hat{\Ab}_F,\hat{\Bb}_F)$ corresponding to $\hat{\lambda}_j(F)$ such that 
\[
\bxi_j^T \hat{\Bb}_F\bxi_k = \begin{cases}
1 & \mathrm{if\; } j=k,\\
0 & \mathrm{if\; } j\ne k.
\end{cases}
\]
Assume that $\tilde{\vb} =\sum_{j=1}^{k'} \alpha_j \bxi_j$ and by definition we have $\yb(F) = \bxi_1 / \|\bxi_1\|_2$.  By assumption, we have $\yb(F)^T\tilde{\vb} = 1-\delta$.  
This implies that $\|\yb(F)-\tilde{\vb}\|_2^2 = 2\delta$.
Also, note that 
\begin{equation*}
\begin{split}
\|\tilde{\vb} -\yb(F)\|_{\hat{\Bb}_F}^2 &=\|\tilde{\vb} -\alpha_1 \bxi_1- (\yb(F)- \alpha_1 \bxi_1)\|_{\hat{\Bb}_F}^2\\ 
&=  \|\tilde{\vb} - \alpha_1 \bxi_1\|_{\hat{\Bb}_F}^2+\|\yb(F) - \alpha_1 \bxi_1\|_{\hat{\Bb}_F}^2-  2 [\yb(F)-\alpha_1\bxi_1]^T \hat{\Bb}_F(\tilde{\vb}-\alpha_1\bxi_1)
\end{split}
\end{equation*}
Since $\yb(F)-\alpha_1 \bxi_1$ is orthogonal to $\tilde{\vb} - \alpha_1 \bxi_1$ under the normalization of $\hat{\Bb}_F$, we have
\begin{equation} 
\label{Eq:lemma:keyproof1}
\sum_{j=2}^{k'} \alpha_j^2 = \|\tilde{\vb} -\alpha_1 \bxi_1\|_{\hat{\Bb}_F}^2 \le \|\tilde{\vb} -\yb(F)\|_{\hat{\Bb}_F}^2 \le 2  \lambda_{\max} (\hat{\Bb}_F) \delta,
\end{equation}
in which the last inequality holds by an application of H\"older's inequality and the fact that $\|\yb(F)-\tilde{\vb}\|_2^2 = 2\delta$.
Moreover, we have  
\begin{equation}
\label{Eq:lemma:keyproof2}
\sum_{j=1}^{k'} \alpha_j^2 = \|\tilde{\vb}\|_{\hat{\Bb}_F}^2  \ge \lambda_{\max} (\hat{\Bb}_F) / \hat{\kappa} \qquad \mathrm{and} \qquad \alpha_1^2 \ge  \lambda_{\max} (\hat{\Bb}_F)/\hat{\kappa}- \sum_{j=2}^{k'} \alpha_j^2  \ge \frac{2\lambda_{\max} (\hat{\Bb}_F)}{3\hat{\kappa} } ,
\end{equation}
where the last inequality is obtained by (\ref{Eq:lemma:keyproof1}) and the assumption that $\delta \le 1/(8c_{\mathrm{upper}}\kappa)$.

We also need a lower bound on $\|\yb(F)\|_{\hat{\Bb}_F}$.  By the triangle inequality, we have
\begin{equation}
\begin{split}
\label{Eq:lemma:keyproof3}
\|\yb(F)\|_{\hat{\Bb}_F} &\ge  \|\tilde{\vb} \|_{\hat{\Bb}_F}- \| \tilde{\vb} -\yb(F)\|_{\hat{\Bb}_F}\ge \sqrt{\sum_{j=1}^{k'} \alpha_j^2} - \sqrt{\lambda_{\max} (\hat{\Bb}_F)} \cdot \|\tilde{\vb}-\yb(F)\|_2
\\
&\ge \frac{1}{2}\sqrt{\sum_{j=1}^{k'} \alpha_j^2} + \frac{1}{2} \sqrt{\frac{\lambda_{\max} (\hat{\Bb}_F)}{\hat{\kappa}}} - \sqrt{2\lambda_{\max}(\hat{\Bb}_F) \delta}
\ge \frac{1}{2} \alpha_1,
\end{split}
\end{equation}
where the second inequality holds by the definition of $\|\tilde{\vb}\|_{\hat{\Bb}_F}$ and an application of H\"older's inequality, the third inequality follows from (\ref{Eq:lemma:keyproof2}), and the last inequality follows from the fact that 
$1/2 \cdot \sqrt{\lambda_{\max}(\hat{\Bb}_F) / \hat{\kappa}} \ge \sqrt{2\lambda_{\max}(\hat{\Bb}_F) \delta}$ under the assumption that $1/(8c_{\mathrm{upper}}\kappa)$.\\

\textbf{Lower and upper bounds for $[\hat{\lambda}_1(F)-\rho]/\rho$:}
To obtain a lower bound for the quantity $\yb(F)^T\vb'$, we need both lower bound and upper bound for the quantity $[\hat{\lambda}_1(F)-\rho]/\rho$.  Recall that $\rho = \tilde{\vb}^T \hat{\Ab}_F \tilde{\vb}/ \tilde{\vb}^T \hat{\Bb}_F \tilde{\vb}$.  Using the fact that 
$ \tilde{\vb}^T \hat{\Ab}_F \tilde{\vb} = \sum_{j=1}^{k'} \alpha_j^2 \hat{\lambda}_j (F)$, we obtain
\begin{equation}
\label{Eq:lemma:keyproof4}
\frac{\hat{\lambda}_1 (F) - \rho}{\rho}  = \frac{\sum_{j=1}^{k'} [\hat{\lambda}_1(F) - \hat{\lambda}_j (F) ]\alpha_j^2 }{\sum_{j=1}^{k'} \hat{\lambda}_j (F)  \alpha_j^2} \le \frac{\hat{\lambda}_1 (F)  \sum_{j=2}^{k'} \alpha_j^2}{\hat{\lambda}_1 (F) \alpha_1^2 } \le \frac{2 \lambda_{\max} (\hat{\Bb}_F) \delta }{\alpha_1^2} \le 3\delta \hat{\kappa},
\end{equation}
where the second to the last inequality holds by (\ref{Eq:lemma:keyproof1}) and the last inequality holds by (\ref{Eq:lemma:keyproof2}).
We now establish a lower bound for $[\hat{\lambda}_1(F)-\rho]/\rho$.  
First, we observe that
\begin{equation}
\label{Eq:lemma:keyproof5}
\delta \le 2\delta - \delta^2  =  (1-\delta)^2 + 1 - 2(1-\delta)\yb(F)^T \tilde{\vb}   =  \|\tilde{\vb} -(1-\delta) \yb(F)\|_2^2 \le \|\tilde{\vb} - \alpha_1 \bxi_1\|_2^2, 
\end{equation}
where the first equality follows from the fact that $\yb(F)^T \tilde{\vb} = 1-\delta$, and the second inequality holds by the fact that $(1-\delta) \yb(F)$ is the scalar projection of $\yb(F)$ onto the vector $\bxi_1$.
Thus, we have 
\begin{equation}
\label{Eq:lemma:keyproof6}
\begin{split}
&\frac{\hat{\lambda}_1 (F) - \rho}{\rho} =\frac{\sum_{j=1}^{k'} [\hat{\lambda}_1(F) - \hat{\lambda}_j (F) ]\alpha_j^2 }{\sum_{j=1}^{k'} \hat{\lambda}_j (F) \alpha_j^2} \ge \frac{[\hat{\lambda}_1(F) - \hat{\lambda}_2 (F)]\sum_{j=2}^{k'} \alpha_j^2}{\hat{\lambda}_1 (F) \alpha_1^2 + \hat{\lambda}_2 (F) \sum_{j=2}^{k'} \alpha_j^2 }\\
&\quad=  \frac{[\hat{\lambda}_1(F) - \hat{\lambda}_2 (F)]\cdot \|\tilde{\vb} - \alpha_1 \bxi_1\|_{\hat{\Bb}_F}^2}{\hat{\lambda}_1 (F) \alpha_1^2 + \hat{\lambda}_2 (F)  \cdot \|\tilde{\vb} - \alpha_1 \bxi_1\|_{\hat{\Bb}_F}^2 }\ge \frac{(1-\gamma) \cdot [\lambda_{\max}(\hat{\Bb}_F)/\hat{\kappa}] \cdot \|\tilde{\vb} - \alpha_1 \bxi_1\|_2^2}{\alpha_1^2 + \gamma \cdot [\lambda_{\max}(\hat{\Bb}_F)/\hat{\kappa}]\cdot \|\tilde{\vb} - \alpha_1 \bxi_1\|_{2}^2}\\
&\quad \ge \frac{(1-\gamma) \cdot \lambda_{\max}(\hat{\Bb}_F) \cdot \delta}{\alpha_1^2\cdot \hat{\kappa} + \gamma \cdot \lambda_{\max}(\hat{\Bb}_F)\cdot \delta},
\end{split}
\end{equation}
where the second to the last inequality holds by dividing the numerator and denominator by $\hat{\lambda}_1(F)$ and using the upper bound $\hat{\lambda}_2(F)/\hat{\lambda}_1(F) \le \gamma$, and the last inequality holds by (\ref{Eq:lemma:keyproof5}).\\

\textbf{Lower bound for $\|\Cb_F \tilde{\vb}\|_2^{-1}$:} In the sequel, we first establish an upper bound for $\|\Cb_F \tilde{\vb}\|_2^2$.  By the definition that $\rho = \tilde{\vb}^T \hat{\Ab}_F \tilde{\vb} / \tilde{\vb}^T \hat{\Bb}_F \tilde{\vb}$, we have 
\$
\tilde{\vb}^T \hat{\Ab}_F \tilde{\vb} - \rho \tilde{\vb}^T \hat{\Bb}_F \tilde{\vb} = 0.
\$  
Moreover, by the definition of $\tilde{\vb} = \sum_{j=1}^{k'} \alpha_j \bxi_j$ and the fact that $\hat{\Ab}_F \bxi_j = \hat{\lambda}_j(F) \hat{\Bb}_F \bxi_j$, we have 
\begin{equation}
\label{Eq:lemma:keyproof7}
\begin{split}
\|(\hat{\Ab}_F-\rho \hat{\Bb}_F) \tilde{\vb}\|_2^2 = \left\|\sum_{j=1}^{k'} \alpha_j \hat{\Ab}_F \bxi_j   - \rho \sum_{j=1}^{k'} \alpha_j \hat{\Bb}_F \bxi_j  \right\|_2^2 = \left\|\sum_{j=1}^{k'} \alpha_j [\hat{\lambda}_j (F) - \rho] \hat{\Bb}_F \bxi_j \right\|_2^2.
\end{split}
\end{equation}
Thus, by (\ref{Eq:lemma:keyproof7}) and the fact that $\tilde{\vb}^T \hat{\Ab}_F \tilde{\vb} - \rho \tilde{\vb}^T \hat{\Bb}_F \tilde{\vb} = 0$, we obtain
\begin{equation}
\label{Eq:lemma:keyproof8}
\|\Cb_F \tilde{\vb}\|_2^2 = \left\| \left[\Ib + \frac{\eta}{\rho} (\hat{\Ab}_F-\rho \hat{\Bb}_F) \right]\tilde{\vb}   \right\|_2^2 =1 +\left\|\sum_{j=1}^{k'} \alpha_j \cdot \left( \frac{\eta}{\rho} \right)\cdot [\hat{\lambda}_j (F) - \rho] \cdot \hat{\Bb}_F \bxi_j \right\|_2^2.
\end{equation}
It remains to establish an upper bound for the second term in the above equation.
Note that by the assumption that $ \delta \le 1/(3\cdot c_{\mathrm{upper}} \kappa)\cdot(1/\gamma-1)$ and (\ref{Eq:lemma:keyproof4}), we have 
\$
\hat{\lambda}_2 (F) \le \rho \le \hat{\lambda}_1 (F).
\$ Moreover, since $\|\tilde{\vb}\|_2^2=1$, we have $\alpha_1^2 \le \lambda_{\max} (\hat{\Bb}_F)$.
Thus, 
\begin{equation}
\label{Eq:lemma:keyproof9}
\begin{split}
&\left\|\sum_{j=1}^{k'} \alpha_j \cdot \left( \frac{\eta}{\rho} \right)\cdot [\hat{\lambda}_j (F) - \rho] \cdot \hat{\Bb}_F \bxi_j \right\|_2^2\\
&\quad \le \alpha_1^2 (\hat{\lambda}_1(F) -\rho)^2 \lambda_{\max}(\hat{\Bb}_F) \cdot (\eta / \rho)^2 + \lambda_{\max} (\hat{\Bb}_F) \sum_{j=2}^{k'} \alpha_j^2 \cdot (\eta/\rho)^2  [\hat{\lambda}_{j}(F)-\rho]^2\\
&\quad \le \lambda_{\max}^2 (\hat{\Bb}_F)  \cdot \eta^2 \cdot  (3\delta \hat{\kappa})^2 +  \lambda_{\max} (\hat{\Bb}_F) \cdot \eta^2 \cdot [{\hat{\lambda}_1 (F)/\rho-1}]^2\cdot \sum_{j=2}^{k'} \alpha_j^2 \\
&\quad \le \lambda_{\max}^2 (\hat{\Bb}_F)  \cdot \eta^2 \cdot  (3\delta \hat{\kappa})^2 + 2 \lambda_{\max}^2 (\hat{\Bb}_F) \cdot \eta^2 \cdot \delta \cdot(3\delta \hat{\kappa})^2\\
&\quad =  9 \cdot \lambda_{\max}^2 (\hat{\Bb}_F)  \cdot \eta^2 \cdot  \delta^2 \cdot \hat{\kappa}^2+   18 \cdot \lambda_{\max}^2 (\hat{\Bb}_F)  \cdot \eta^2 \cdot  \delta^3 \cdot \hat{\kappa}^2,
\end{split}
\end{equation}
where the second inequality is from (\ref{Eq:lemma:keyproof4}) and the third inequality follows from (\ref{Eq:lemma:keyproof1}).
Substituting (\ref{Eq:lemma:keyproof9}) into (\ref{Eq:lemma:keyproof8}),  we have 
\begin{equation}
\label{Eq:lemma:keyproof10}
\begin{split}
\|\Cb_F \tilde{\vb}\|_2^2 &\le 1 + 9 \cdot \lambda_{\max}^2 (\hat{\Bb}_F)  \cdot \eta^2 \cdot  \delta^2 \cdot \hat{\kappa}^2+   18 \cdot \lambda_{\max}^2 (\hat{\Bb}_F)  \cdot \eta^2 \cdot  \delta^3 \cdot \hat{\kappa}^2\\
&\le 1 + 12 \cdot \lambda_{\max}^2 (\hat{\Bb}_F)  \cdot \eta^2 \cdot  \delta^2 \cdot \hat{\kappa}^2,
\end{split}
\end{equation}
where the last inequality follows from the fact that $2\delta \le 1/4$, which holds by the assumption that $\delta \le 1/(8c_{\mathrm{upper}} \kappa)$. Meanwhile, note that the second term in the upper bound is less than one by the assumption $\delta \le 1/(8c_{\mathrm{upper}} \kappa)$ and $\eta c_{\mathrm{upper}} \lambda_{\max} (\Bb)<1$.
Hence, by invoking (\ref{Eq:lemma:keyproof10}) ~nd the fact that $1/\sqrt{1+y} \ge 1-y/2$ for $|y| <1$, we have 
\begin{equation}
\label{Eq:lemma:keyproof11}
\|\Cb_F \tilde{\vb}\|_2^{-1}\ge 1 - 6 \cdot \lambda_{\max}^2 (\hat{\Bb}_F)  \cdot \eta^2 \cdot  \delta^2 \cdot \hat{\kappa}^2.
\end{equation}
\\
\textbf{Lower bound for $\yb(F)^T \Cb_F \tilde{\vb}$:} We have 
\begin{equation}
\label{Eq:lemma:keyproof12}
\begin{split}
\yb(F)^T \Cb_F \tilde{\vb} &= \yb(F)^T \tilde{\vb} + \frac{\eta}{\rho} \cdot  \yb(F)^T (\hat{\Ab}_F-\rho \hat{\Bb}_F) \tilde{\vb}\\
&= 1-\delta + \frac{\eta}{\rho }\cdot [\hat{\lambda}_1 (F) - \rho ]\cdot \yb(F)^T \hat{\Bb}_F \tilde{\vb}\\
&=1-\delta + \frac{\eta}{\rho }\cdot [\hat{\lambda}_1 (F) - \rho ]\cdot  \biggl( \alpha_1 \cdot \frac{ \bxi_1^T \hat{\Bb}_F \bxi_1}{\|\bxi_1\|_2}  \biggr)\\
&=1-\delta + \eta \cdot \alpha_1 \cdot \biggl[\frac{\hat{\lambda}_1 (F)  - \rho}{\rho} \biggr]\cdot  \|\yb(F)\|_{\hat{\Bb}_F}\\
&\ge 1-\delta  + \frac{1}{2}\cdot  \eta \cdot \alpha_1^2 \cdot   \biggl[ \frac{(1-\gamma) \cdot \lambda_{\max}(\hat{\Bb}_F) \cdot \delta}{\alpha_1^2\cdot \hat{\kappa} + \gamma \cdot \lambda_{\max}(\hat{\Bb}_F)\cdot \delta}
 \biggr] \\
 &\ge 1-\delta +  \frac{1}{2}\cdot \eta \cdot \frac{\alpha_1^2 \cdot (1-\gamma) \cdot \delta}{\hat{\kappa} + \gamma}\\
  &\ge 1-\delta +  \frac{1}{3}\cdot \eta \cdot \lambda_{\min}(\hat{\Bb}_F)\cdot \frac{ (1-\gamma) \cdot \delta}{   (\hat{\kappa} + \gamma)},
 \end{split}
\end{equation}
where the first inequality follows from (\ref{Eq:lemma:keyproof3}) and (\ref{Eq:lemma:keyproof6}), the second inequality uses the fact that $\alpha_1^2 \le \lambda_{\max} (\hat{\Bb}_F)$, and the last inequality follows from (\ref{Eq:lemma:keyproof2}).\\

\textbf{Combining the results:} We now establish a lower bound on $\yb(F)^T \vb'$. 
From (\ref{Eq:lemma:keyproof11}) and (\ref{Eq:lemma:keyproof12}), we have 
\begin{equation}
\label{Eq:lemma:keyproof13}
\begin{split}
\yb(F)^T \vb' &= \yb(F)^T  \Cb_F \tilde{\vb} \cdot \|\Cb_F \tilde{\vb}\|^{-1}_2\\
&\ge \left(1-\delta + \frac{1}{3} \cdot \eta \cdot \lambda_{\min}(\hat{\Bb}_F) \cdot \left[   
 \frac{(1-\gamma)\cdot  \delta}{ (\hat{\kappa}+ \gamma) }
 \right]  \right) \cdot \left( 1 - 6  \cdot  \lambda_{\max}^2 (\hat{\Bb}_F) \cdot \eta^2\cdot  \delta^2 \cdot \hat{\kappa}^2 \right)\\
 &\ge 1-\delta + \frac{1}{3} \cdot \eta\cdot  \lambda_{\min} (\hat{\Bb}_F)\cdot  \left[ \frac{(1-\gamma)\cdot \delta}{ (\hat{\kappa}+\gamma)}\right] - 6    \cdot \lambda_{\max}^2 (\hat{\Bb}_F)\cdot  \eta^2\cdot  \delta^2\cdot  \hat{\kappa}^2\\
 &\quad  - 2 \cdot  \hat{\kappa}^2\cdot  \eta^3 \cdot    \lambda_{\max}^3 (\hat{\Bb}_F) \cdot   \delta^2\cdot  \left[ \frac{(1-\gamma)\cdot \delta}{ (\hat{\kappa}+\gamma) }\right]\\
  &\ge 1-\delta + \frac{1}{3} \cdot \eta\cdot  \lambda_{\min} (\hat{\Bb}_F)\cdot  \left[ \frac{(1-\gamma)\cdot \delta}{ (\hat{\kappa}+\gamma)}\right] - 6.25 \cdot \lambda_{\max}^2 (\hat{\Bb}_F)\cdot  \eta^2\cdot  \delta^2\cdot  \hat{\kappa}^2\\
 &\ge 1-\delta + \frac{1}{8}\cdot \eta\cdot \lambda_{\min} (\hat{\Bb}_F) \cdot \left[ \frac{(1-\gamma) \delta
}{(\hat{\kappa} + \gamma)}\right],
\end{split}
\end{equation}
in which the third inequality holds by the assumption that the step size $\eta$ is sufficiently small such that $\eta \lambda_{\max} (\hat{\Bb}_F)<1$, and the  last inequality holds under the condition that 
\[
\frac{1-\gamma}{ (\hat{\kappa}+\gamma)} \ge 30 \eta \lambda_{\max} (\hat{\Bb}) \delta \hat{\kappa}^2,
\]
which is implied by the following inequality under Assumption~\ref{ass:large n}
\[
\delta \le  
 \frac{1-\gamma}{30\cdot (1+c) \cdot c_{\mathrm{upper}}^2\cdot \eta \cdot \lambda_{\max} (\Bb) \cdot \kappa^2 \cdot (c_{\mathrm{upper}}\kappa+\gamma)}.
\]
By Assumption~\ref{ass:large n}, we have 
\[
\yb(F)^T \vb' \ge 1-\delta + \frac{1+c}{8}\cdot \eta\cdot  \lambda_{\min} ({\Bb})\cdot  [1-\yb(F)^T \tilde{\vb}]\cdot  \left( \frac{1-\gamma}{  c_{\mathrm{upper}}{\kappa}+\gamma}\right),
\]
as desired.

\end{proof}

\subsection{Proof of Lemma~\ref{lemma:perturbation vec}}
\begin{proof}
 The first part of the lemma on the following inequality follows directly from Theorem 4.3 in \citet{stewart1979pertubation}
\[
\frac{\|\vb(F) -\vb^*\|_2}{\|\vb^*\|_2}\le \frac{\delta(F)}{ \Delta \hat{\lambda} \cdot \mathrm{cr}(\hat{\Ab}_F,\hat{\Bb}_F) }.
\]
We now prove the second part of the lemma.  

Setting $\yb(F) = \vb(F)/\|\vb(F)\|_2$ and $\yb^* = {\vb^*}/\|{\vb^*}\|_2$ such that $\|\yb(F)\|_2= 1 $ and $\|{\yb^*}\|_2=1$, we have
\begin{equation*}
\begin{split}
\|\yb(F) - {\yb}^* \|_2&\le \left\| \frac{\vb(F)}{\|\vb (F)\|_2}  - \frac{{\vb^*}}{\|{\vb^*}\|_2}       \right\|_2   \\
&\le \frac{1}{\|\vb(F)\|_2 \cdot\|{\vb^*}\|_2}  \cdot \|  \vb(F)\cdot \|{\vb^*}\|_2 - {\vb^*} \cdot \|\vb(F)\|_2     \|_2\\
&\le \frac{2}{\|\vb^*\|_2}  \cdot \|  \vb(F) - {\vb^*}     \|_2  \\
&\le  2  \frac{\delta(F)}{ \Delta \hat{\lambda} \cdot \mathrm{cr}(\hat{\Ab}_F,\hat{\Bb}_F) }
\end{split}
\end{equation*}
where the third inequality holds by adding and subtracting $\vb(F) \cdot \|\vb (F)\|_2$.
By definition,  $\delta(F) \le \epsilon(k')$ and $\Delta \hat{\lambda} \ge \Delta \lambda$. Moreover, by Theorem 2.4 in \citet{stewart1979pertubation}, $\mathrm{cr}(\hat{\Ab}_F,\hat{\Bb}_F)\ge \mathrm{cr}(k')-\epsilon(k')$.  Thus, we obtain 
\[
\|\yb(F)-\yb^*\|_2 \le \frac{2}{\Delta \lambda \cdot (\mathrm{cr}(k')-\epsilon(k'))} \cdot \epsilon(k').
\]
The other case for $\|\yb(F)+\yb^*\|_2$ can be proven similarly.
\end{proof}

\section{Proof of Proposition~\ref{prop:initial}}
\label{proof theorem 2}

\begin{proof}
The proof is an adaptation of the proof of Theorem 4.1 in \citet{gao2014sparse} and the proof of Theorem 1 in \citet{SIR2018}, with some modifications to the curvature lemma to remove the structural assumptions on $\Ab$.  Without loss of generality, we assume that $\Ab$ is full rank. 
For ease of notation, throughout the proof, we write $\Vb$, $\bLambda$, $\Pb$ to indicate $\Vb^*$, $\bLambda^*$, and $\Pb^*$, respectively.

Let $\Vb = (\Vb_{\cdot K},\Vb_{\cdot K^{c}}) \in \RR^{d\times d}$,
where $\Vb_{\cdot K} \in \RR^{d\times K}$ are the $K$ leading generalized eigenvectors of $(\Ab,\Bb)$  and $\Vb_{\cdot K^c}\in\RR^{d\times (d-K)}$ are the last $d-K$ generalized eigenvectors.   Let $\bLambda\in \RR^{d\times d}$ be a diagonal matrix of the generalized eigenvalues.  Let $\cS_v$ be a set containing indices of non-zero rows of $\Vb\in \RR^{d\times d}$, with cardinality $|\cS_v| =s$. 
In other words, each generalized eigenvector has at most $s$ nonzero elements.
Let $\Pb = \Vb_{\cdot K} \Vb_{\cdot K}^T$ and let $\cS$ and $\cS^c$ be the support of $\Pb$ and complimentary set of $\cS$, respectively. 
To facilitate the proof, we define some new notation 
\[
\tilde{\Ab} = \hat{\Bb} \Vb \bLambda \Vb^T \hat{\Bb}, \qquad \tilde{\Vb}_{\cdot K} = \Vb_{\cdot K} (\Vb_{\cdot K}^T \hat{\Bb} \Vb_{\cdot K} )^{-1/2},\qquad \tilde{\Pb} = \tilde{\Vb}_{\cdot K} \tilde{\Vb}_{\cdot K}^T.
\]
Let $\hat{\Pb}$ be a solution of~\eqref{eq:convexrelaxation} with tuning parameter $K$ and $\zeta$, and let $\bDelta = \hat{\Pb}-\tilde{\Pb}$.
Finally, for two matrices $\Eb$ and $\Fb$, we write $\langle \Eb,\Fb \rangle = \mathrm{tr}(\Eb\Fb)$.

It can be shown that $\tilde{\Pb}$ satisfies both constraints in~\eqref{eq:convexrelaxation}, and therefore is a feasible solution of~\eqref{eq:convexrelaxation}.  
Since $\tilde{\Pb}$ is a feasible solution of~\eqref{eq:convexrelaxation} and $\hat{\Pb}$ is the optimal solution of \eqref{eq:convexrelaxation}, we have 
\[
-\langle \hat{\Ab},\hat{\Pb} \rangle + \zeta \|\hat{\Pb}\|_{1,1}\le -\langle \hat{\Ab},\tilde{\Pb} \rangle + \zeta \|\tilde{\Pb}\|_{1,1}.
\]
By picking $\zeta > 2 \|\hat{\Ab}-\tilde{\Ab}\|_{\infty,\infty} $, triangle inequality, rearranging the terms, and using the fact that $\tilde{\Pb}$ and $\Pb$ share the same support, it can be shown that 
\begin{equation}
\label{eq:crp1}
-\langle \tilde{\Ab},\bDelta \rangle \le \frac{3\zeta}{2} \|\bDelta_{\cS}\|_{1,1} - \frac{\zeta}{2} \|\bDelta_{\cS^c}\|_{1,1}.
\end{equation}

The main difference between our proof and that of \citet{gao2014sparse} and \citet{SIR2018} is in obtaining the lower bound for $-\langle \tilde{\Ab},\bDelta \rangle $.  
By the definition of $\tilde{\Ab}, \tilde{\Pb}$, we obtain 
\begin{equation}
\label{eq:crp2}
\begin{split}
&-\langle \tilde{\Ab},\bDelta \rangle\\
& = \langle \hat{\Bb} \Vb \bLambda\Vb^T\hat{\Bb},\tilde{\Pb}-\hat{\Pb} \rangle\\
&=   \langle \hat{\Bb}^{1/2} \Vb \bLambda\Vb^T\hat{\Bb}^{1/2},\hat{\Bb}^{1/2}(\tilde{\Pb}-\hat{\Pb})\hat{\Bb}^{1/2} \rangle\\
&=   \langle \hat{\Bb}^{1/2} \Vb \bLambda\Vb^T\hat{\Bb} \tilde{\Pb} \hat{\Bb}^{1/2}, \Ib  - \hat{\Bb}^{1/2} \hat{\Pb} \hat{\Bb}^{1/2} \rangle
- \langle    
(\Ib- \hat{\Bb}^{1/2} \tilde{\Pb} \hat{\Bb}^{1/2}) \hat{\Bb}^{1/2} \Vb \bLambda\Vb^T\hat{\Bb}^{1/2}, \hat{\Bb}^{1/2} \hat{\Pb} \hat{\Bb}^{1/2}
\rangle \\
&= I - II.
\end{split}
\end{equation}
It suffices to obtain a lower bound for $I$ and an upper bound for $II$.

\textbf{Lower bound for $I$:}
We have 
\begin{equation}
\label{eq:crp2-2}
\begin{split}
I &= \mathrm{tr}[\hat{\Bb}^{1/2} \Vb \bLambda\Vb^T\hat{\Bb} \Vb_{\cdot K} (\Vb_{\cdot K}^T \hat{\Bb}\Vb_{\cdot K})^{-1} \Vb_{\cdot K}^T  \hat{\Bb}^{1/2}(\Ib  - \hat{\Bb}^{1/2} \hat{\Pb} \hat{\Bb}^{1/2}) ]\\
&=\mathrm{tr}[\hat{\Bb}_{\cS_v} \Vb_{\cS_v,K} (\Vb_{\cdot K}^T \hat{\Bb}\Vb_{\cdot K})^{-1} \Vb_{\cdot K}^T  \hat{\Bb}^{1/2}(\Ib  - \hat{\Bb}^{1/2} \hat{\Pb} \hat{\Bb}^{1/2}) \hat{\Bb}^{1/2} \Vb \bLambda\Vb^T_{\cS_v \cdot}]\\
&\ge \frac{\lambda_{\min} (\hat{\Bb}_{\cS_v})}{\lambda_{\max} (\Bb_{\cS_v})} \cdot 
\mathrm{tr}[\hat{\Bb}^{1/2} \Vb \bLambda\Vb^T_{\cS_v\cdot}\Bb_{\cS_v} \Vb_{\cS_v,K} (\Vb_{\cdot K}^T \hat{\Bb}\Vb_{\cdot K})^{-1} \Vb_{\cdot K}^T  \hat{\Bb}^{1/2}(\Ib  - \hat{\Bb}^{1/2} \hat{\Pb} \hat{\Bb}^{1/2}) ]\\
&\ge \frac{\lambda_{\min} ({\Bb}_{\cS_v}) - \rho(\Eb_{\Bb},s)}{\lambda_{\max} (\Bb_{\cS_v})} \cdot 
\mathrm{tr}[\hat{\Bb}^{1/2} \Vb \bLambda\Vb^T_{\cS_v \cdot}\Bb_{\cS_v} \Vb_{\cS_v,K} (\Vb_{\cdot K}^T \hat{\Bb}\Vb_{\cdot K})^{-1} \Vb_{\cdot K}^T  \hat{\Bb}^{1/2}(\Ib  - \hat{\Bb}^{1/2} \hat{\Pb} \hat{\Bb}^{1/2}) ]\\
&\ge \frac{(1-c)}{\kappa(\Bb)}\cdot 
\mathrm{tr}[\hat{\Bb}^{1/2} \Vb \bLambda\Vb^T_{\cS_v \cdot}\Bb_{\cS_v} \Vb_{\cS_v,K} (\Vb_{\cdot K}^T \hat{\Bb}\Vb_{\cdot K})^{-1} \Vb_{\cdot K}^T  \hat{\Bb}^{1/2}(\Ib  - \hat{\Bb}^{1/2} \hat{\Pb} \hat{\Bb}^{1/2}) ],
\end{split}
\end{equation}
where the second inequality holds by Weyl's inequality, i.e., $\lambda_{\min} (\Bb_{\cS_v}) \le \lambda_{\min} (\hat{\Bb}_{\cS_v})  + \rho(\Eb_\Bb,s)$, and the last inequality follows from Assumption~\ref{ass:large n}.
Note that
\begin{equation*}
\begin{split}
&\mathrm{tr}[\hat{\Bb}^{1/2} \Vb \bLambda\Vb^T_{\cS_v \cdot}\Bb_{\cS_v} \Vb_{\cS_v,K} (\Vb_{\cdot K}^T \hat{\Bb}\Vb_{\cdot K})^{-1} \Vb_{\cdot K}^T  \hat{\Bb}^{1/2}(\Ib  - \hat{\Bb}^{1/2} \hat{\Pb} \hat{\Bb}^{1/2}) ] \\
&= \mathrm{tr} \left[ 
\hat{\Bb}^{1/2} (\Vb_{\cdot K},\Vb_{\cdot K^c}) 
\begin{pmatrix}
\bLambda_K & \mathbf{0}\\
\mathbf{0} & \bLambda_{K^c}
\end{pmatrix}
\begin{pmatrix}
\Ib_K \\ \mathbf{0}
\end{pmatrix}
 (\Vb_{\cdot K}^T \hat{\Bb}\Vb_{\cdot K})^{-1} \Vb_{\cdot K}^T  \hat{\Bb}^{1/2}(\Ib  - \hat{\Bb}^{1/2} \hat{\Pb} \hat{\Bb}^{1/2})
\right]\\ 
&=\mathrm{tr} \left[ 
\hat{\Bb}^{1/2} \Vb_{\cdot K} \bLambda_K  (\Vb_{\cdot K}^T \hat{\Bb}\Vb_{\cdot K})^{-1} \Vb_{\cdot K}^T  \hat{\Bb}^{1/2}(\Ib  - \hat{\Bb}^{1/2} \hat{\Pb} \hat{\Bb}^{1/2})
\right]. 
\end{split}
\end{equation*}
Substituting this into~\eqref{eq:crp2-2}, we obtain 
\begin{equation}
\label{eq:crp2-3}
I \ge \frac{(1-c)\lambda_K}{ \kappa(\Bb)} \cdot \langle
\hat{\Bb}^{1/2} \tilde{\Pb} \hat{\Bb}^{1/2},\Ib  - \hat{\Bb}^{1/2} \hat{\Pb} \hat{\Bb}^{1/2}
\rangle. 
\end{equation}

\textbf{Upper bound for $II$:}  Observe that 
\begin{equation*}
\begin{split}
&(\Ib- \hat{\Bb}^{1/2} \tilde{\Pb} \hat{\Bb}^{1/2}) \hat{\Bb}^{1/2} \Vb \bLambda\Vb^T\hat{\Bb}^{1/2} \\
&= \hat{\Bb}^{1/2} \Vb_{\cdot K} \bLambda_{K}\Vb^T_{\cdot K}\hat{\Bb}^{1/2}+\hat{\Bb}^{1/2} \Vb_{\cdot K^c} \bLambda_{K^c}\Vb^T_{ \cdot K^c}\hat{\Bb}^{1/2}
- \hat{\Bb}^{1/2} \Vb_{\cdot K} \bLambda_{K}\Vb^T_{\cdot K}\hat{\Bb}^{1/2}\\
&\quad - \hat{\Bb}^{1/2} \Vb_{\cdot K} (\Vb_{\cdot K}^T \hat{\Bb} \Vb_{\cdot K})^{-1} \Vb_{\cdot K}^T  \hat{\Bb} \Vb_{\cdot K^c}\bLambda_{K^c}\Vb^T_{\cdot K^c} \hat{\Bb}^{1/2}\\
&= (\Ib- \hat{\Bb}^{1/2} \tilde{\Pb} \hat{\Bb}^{1/2}) \hat{\Bb}^{1/2} \Vb_{\cdot K^c} \bLambda_{K^c}\Vb^T_{\cdot K^c}\hat{\Bb}^{1/2},
\end{split}
\end{equation*}
where the last equality holds since the first equality depends only on $\bLambda_{K^c}$.
Thus, we have 
\begin{equation}
\label{eq:crp3}
\begin{split}
II &= \langle    
(\Ib- \hat{\Bb}^{1/2} \tilde{\Pb} \hat{\Bb}^{1/2}) \hat{\Bb}^{1/2} \Vb_{\cdot K^c} \bLambda_{K^c}\Vb^T_{\cdot K^c}\hat{\Bb}^{1/2}, \hat{\Bb}^{1/2} \hat{\Pb} \hat{\Bb}^{1/2} \rangle\\
&\le  \lambda_{K+1} \langle    
(\Ib- \hat{\Bb}^{1/2} \tilde{\Pb} \hat{\Bb}^{1/2}) \hat{\Bb}^{1/2} \Vb_{\cdot K^c} \Vb^T_{\cdot K^c}\hat{\Bb}^{1/2}, \hat{\Bb}^{1/2} \hat{\Pb} \hat{\Bb}^{1/2}\rangle\\
&\le  \lambda_{K+1}   \|\Vb^T_{\cdot K^c}\hat{\Bb}\Vb_{\cdot K^c} \|_2 \langle    
\Ib- \hat{\Bb}^{1/2} \tilde{\Pb} \hat{\Bb}^{1/2} , \hat{\Bb}^{1/2} \hat{\Pb} \hat{\Bb}^{1/2}\rangle\\
&\le  \lambda_{K+1} (1+\|\Vb^T_{\cdot K^c}(\hat{\Bb}-\Bb)\Vb_{\cdot K^c} \|_2 ) \langle    
\Ib- \hat{\Bb}^{1/2} \tilde{\Pb} \hat{\Bb}^{1/2} , \hat{\Bb}^{1/2} \hat{\Pb} \hat{\Bb}^{1/2}\rangle,
\end{split}
\end{equation}
where the last inequality holds by adding and subtracting $\Vb^T_{\cdot K^c}\Bb\Vb_{\cdot K^c} $ and the triangle inequality. 
Since only $s$ rows of $\Vb_{\cdot K^c}$ are nonzero, by Holder's inequality, we obtain
\begin{equation*}
\begin{split}
\|\Vb^T_{\cdot K^c}(\hat{\Bb}-\Bb)\Vb_{\cdot K^c} \|_2  &\le \|\Bb^{-1/2}\Bb^{1/2}\Vb_{\cdot K^c}\|_2^2 \cdot \rho(\Eb_{\Bb},s) \\
&\le c \lambda_{\min} (\Bb) \cdot \|\Bb^{-1/2}\|_{2}^2\\
&\le c,
\end{split}
\end{equation*}
where the second inequality holds under Assumption~\ref{ass:large n}.  Substituting this into~\eqref{eq:crp3}, we obtain 
\begin{equation}
\label{eq:crp4}
II \le c \cdot \lambda_{K+1} \cdot  \langle    
\Ib- \hat{\Bb}^{1/2} \tilde{\Pb} \hat{\Bb}^{1/2}, \hat{\Bb}^{1/2} \hat{\Pb} \hat{\Bb}^{1/2}\rangle.
\end{equation}

By definition, $\mathrm{tr} (\hat{\Bb}^{1/2} \hat{\Pb} \hat{\Bb}^{1/2}) =\mathrm{tr} (\hat{\Bb}^{1/2} \tilde{\Pb} \hat{\Bb}^{1/2}) = K$.  Substituting \eqref{eq:crp2-3} and~\eqref{eq:crp4} into \eqref{eq:crp2}, we obtain
\begin{equation}
\label{eq:crp5}
\begin{split}
-\langle  \tilde{\Ab},\bDelta \rangle  &\ge \left[ \frac{(1-c)\lambda_K}{ \kappa(\Bb)} - c\lambda_{K+1}\right]  \left( K - \langle  \hat{\Bb}^{1/2} \hat{\Pb} \hat{\Bb}^{1/2},\hat{\Bb}^{1/2} \tilde{\Pb} \hat{\Bb}^{1/2} \rangle \right)\\
&\ge \frac{1}{2}\left[ \frac{(1-c)\lambda_K}{ \kappa(\Bb)} - c\lambda_{K+1}\right]\cdot \|\hat{\Bb}^{1/2}\bDelta\hat{\Bb}^{1/2}  \|_F^2.
\end{split} 
\end{equation}

The rest of the proof follows from the proof of Theorem 1 in \citet{SIR2018} or the proof of Theorem 4.1 in \citet{gao2014sparse}.  We hereby provide a proof sketch and refer the reader to \citet{SIR2018} for the details. For notational convenient, let $\delta_{\mathrm{gap}} =  [(1-c)\lambda_K / \kappa(\Bb) - c \lambda_{K+1}]$.
Combining \eqref{eq:crp1} and \eqref{eq:crp5}, we have
\begin{equation}
\label{eq:crp6}
\|\hat{\Bb}^{1/2}\bDelta\hat{\Bb}^{1/2}\|_F^2 \le \frac{3\zeta}{\delta_{\mathrm{gap}}} \|\bDelta_{\cS}\|_{1,1}.
\end{equation}
Moreover, $-\langle \tilde{\Ab} , \bDelta \rangle \ge 0$ implies that $\|\bDelta_{\cS^c}\|_{1,1} \le 3 \|\bDelta_{\cS}\|_{1,1}$.  

Similar to \citet{SIR2018}, we partition the set $\cS^c$ into $J$ sets such that $\cS_1^c$ is the index set of the largest $l$ entries in absolute values of $\bDelta$, $\cS_2^c$ is the index the index set of the second largest $l$ entries of $\bDelta$, and so forth, with $|\cS_J^c| \le l$. 
By Lemma S4 of \citet{SIR2018} and the fact that  $\|\bDelta_{\cS^c}\|_{1,1} \le 3 \|\bDelta_{\cS}\|_{1,1}$, we obtain $\sum_{j=2}^J \|\bDelta_{\cS_j^c}\|_F\le 3sl^{-1/2} \|\bDelta_{\cS}\|_F$.
Under Assumption~\ref{ass:large n}, picking $l = c_1 s^2$,  it can be shown that 
\begin{equation}
\label{eq:crp7}
\begin{split}
\|\hat{\Bb}^{1/2}\bDelta \hat{\Bb}^{1/2}\|_F &\ge \|\hat{\Bb}^{1/2}\bDelta_{\cS \cup \cS^c_1} \hat{\Bb}^{1/2}\|_F - \sum_{j=2}^J \|\hat{\Bb}^{1/2}\bDelta_{\cS_j^c} \hat{\Bb}^{1/2}\|_F\\
&\ge \left[(1-c) \lambda_{\min} (\Bb) - \frac{3(1+c)\lambda_{\max} (\Bb)}{c_1}\right]  \|\bDelta_{\cS\cup \cS_1^c}\|_F\\
&\ge C \|\bDelta_{\cS\cup \cS_1^c}\|_F,
\end{split}
\end{equation}
where $C$ is a generic constant, and the last inequality holds by picking $c_1$ to be sufficiently large.

Combining \eqref{eq:crp6} and \eqref{eq:crp7}, 
\[
 \|\bDelta_{\cS\cup \cS_1^c}\|_F \le C \left( \frac{\zeta}{\delta_{\mathrm{gap}}}  \|\bDelta_{\cS }\|_{1,1} \right)^{1/2} \le C \left( \frac{\zeta s}{\delta_{\mathrm{gap}}}  \|\bDelta_{\cS\cup\cS_1^c }\|_F \right)^{1/2}.
\]
By squaring both sides, we obtain $ \|\bDelta_{\cS\cup \cS_1^c}\|_F  \le C \frac{\zeta s}{\delta_{\mathrm{gap}}} $.
By the triangle inequality, 
\begin{equation}
\begin{split}
\|\bDelta\|_F &\le  \|\bDelta_{\cS\cup \cS_1^c}\|_F + \|\bDelta_{(\cS\cup \cS_1^c)^c}\|_F \\ 
&\le   \|\bDelta_{\cS\cup \cS_1^c}\|_F + \sum_{j=2}^J \|\bDelta_{\cS_j^c}\|_F\\
&\le (1+3/c_1)\|\bDelta_{\cS\cup\cS_1^c}\|_F\\
&\le C \frac{\zeta s}{\delta_{\mathrm{gap}}},
\end{split}
\end{equation}
where the second inequality holds by Lemma S4 of \citet{SIR2018} and the third inequality holds by picking $l = c_1 s^2$.
Finally, by the triangle inequality and Lemma S1 of \citet{SIR2018}, we obtain 
\[
\|\hat{\Pb}-\Pb\|_F \le \|\bDelta\|_F + \|\Pb- \tilde{\Pb}\|_F \le C \left( \frac{\zeta s }{\delta_{\mathrm{gap}}} + K\|\hat{\Bb}_{\cS_v}-\Bb_{\cS_v}\|_{2} 
\right).
\]

\end{proof}

\end{document}